%% file: main.tex
\documentclass[10pt,twocolumn,letterpaper]{article}

\usepackage[pagenumbers]{cvpr} 

\usepackage{bm}
\usepackage{float}
\usepackage{multirow}
\usepackage{arydshln} 
\usepackage{booktabs}
\usepackage{algorithm}
\usepackage{algpseudocode}

\input{preamble}
\definecolor{cvprblue}{rgb}{0.21,0.49,0.74}
\usepackage[pagebackref,breaklinks,colorlinks,allcolors=cvprblue]{hyperref}

\def\paperID{14000} 
\def\confName{CVPR}
\def\confYear{2026}

\title{HalluGen: Synthesizing Realistic and Controllable Hallucinations for Evaluating Image Restoration}
\author{
Seunghoi Kim$^{1,2}$\thanks{Correspondence to: seunghoi.kim.17@ucl.ac.uk} \quad
Henry F.~J.~Tregidgo$^{1,2}$ \quad
Chen Jin$^{4}$ \quad
Matteo Figini$^{1,3}$ \quad
Daniel C.~Alexander$^{1,3}$ \\
$^{1}$ Hawkes Institute, UCL \quad
$^{2}$ Dept. of Medical Physics and Biomedical Engineering, UCL \\
$^{3}$ Dept. of Computer Science, UCL \quad
$^{4}$ Centre for AI, DS\&AI, AstraZeneca, UK
}

\begin{document}
\maketitle

\input{sec/0_abstract}    
\input{sec/1_intro}
\input{sec/2_formatting}
\input{sec/3_method1}
\input{sec/6_experiments}
\input{sec/8_conclusion}

{
    \small
    \bibliographystyle{ieeenat_fullname}
    \bibliography{main}
}

\input{sec/X_suppl}

\end{document}

%% file: sec/0_abstract.tex
\begin{abstract}
Generative models are prone to hallucinations: plausible but incorrect structures absent in the ground truth. 
This issue is problematic in image restoration for safety-critical domains such as medical imaging, industrial inspection, and remote sensing, where such errors undermine reliability and trust. 
For example, in low-field MRI, widely used in resource-limited settings, restoration models are essential for enhancing low-quality scans, yet hallucinations can lead to serious diagnostic errors.
Progress has been hindered by a circular dependency: evaluating hallucinations requires labeled data, yet such labels are costly and subjective.
We introduce HalluGen, a diffusion-based framework that synthesizes realistic hallucinations with controllable type, location, and severity, producing perceptually realistic but semantically incorrect outputs (segmentation IoU drops from 0.86 to 0.36).
Using HalluGen, we construct the first large-scale hallucination dataset comprising 4,350 annotated images derived from 1,450 brain MR images for low-field enhancement, enabling systematic evaluation of hallucination detection and mitigation.
We demonstrate its utility in two applications: (1) benchmarking image quality metrics and developing Semantic Hallucination Assessment via Feature Evaluation (SHAFE), a feature-based metric with soft-attention pooling that improves hallucination sensitivity over traditional metrics; and (2) training reference-free hallucination detectors that generalize to real restoration failures.
Together, HalluGen and its open dataset establish the first scalable foundation for evaluating hallucinations in safety-critical image restoration. 
We will make the code and dataset publicly available upon acceptance.

\end{abstract}

%% file: sec/1_intro.tex
\section{Introduction}
\label{sec:intro}

\begin{figure*}[t]
  \centering
  \includegraphics[width=0.95\linewidth]{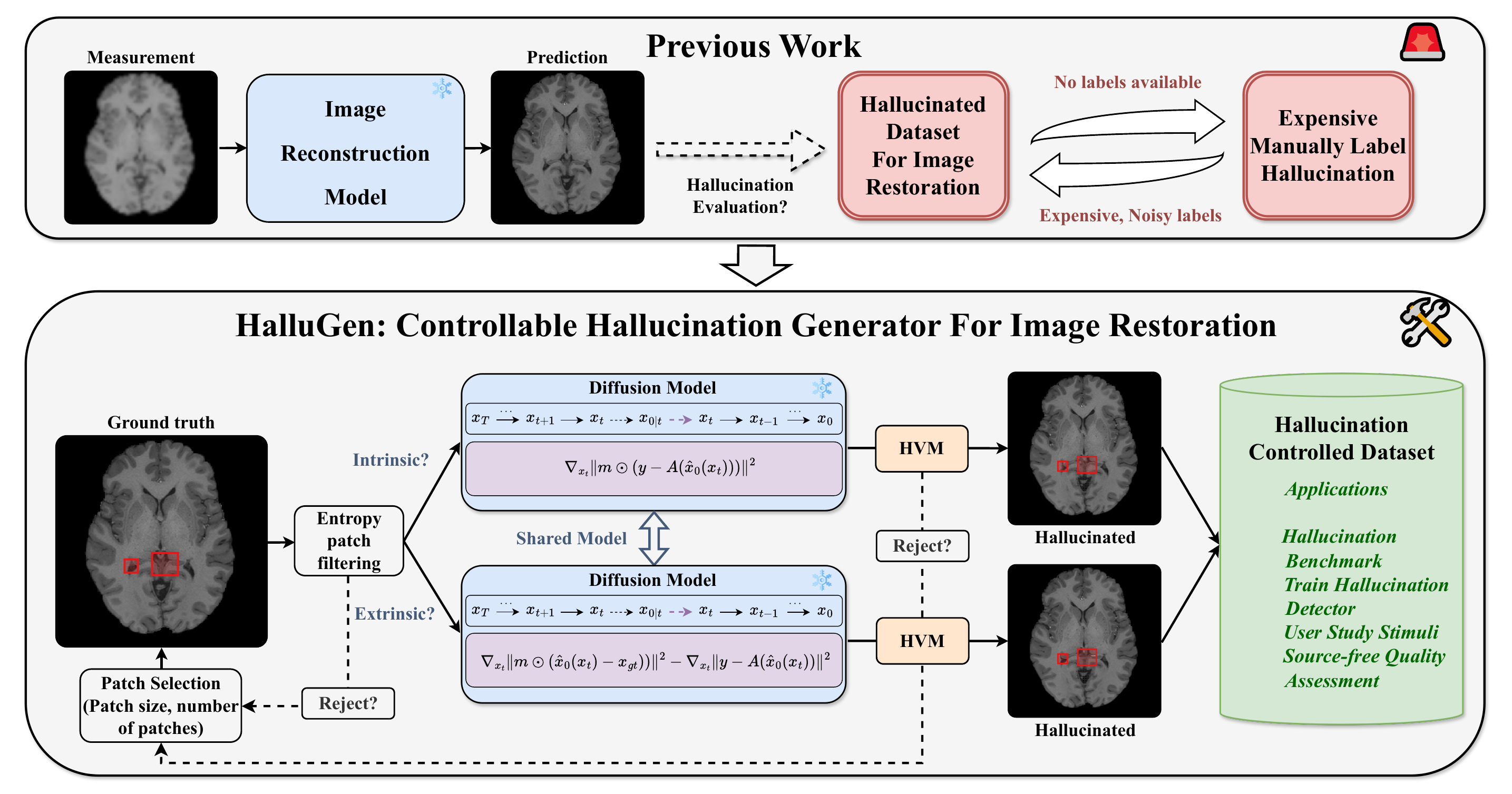}
  \caption{
  \textbf{Circular dependency in hallucination evaluation and our proposed HalluGen solution.} Top: Reliable hallucination analysis requires labeled data, but obtaining labels demands expert annotation with high disagreement.
  Bottom: HalluGen breaks this loop by generating controllable hallucinations with automatic labels, enabling systematic benchmarking, and perceptual studies across domains.
  }
  \label{fig:hallugen}
  \vspace{-12pt} 
\end{figure*}

Hallucinations, visually plausible yet incorrect content, are a fundamental failure mode of modern generative models. In image restoration, where models reconstruct missing information from degraded measurements, such errors can be especially harmful in safety-critical domains such as medical imaging, industrial inspection, and remote sensing. 
Unlike adversarial attacks, hallucinations arise naturally from generative uncertainty. 

Despite recent advances in image restoration~\cite{ESRGAN,edsr,sr_transformer,swinir,deblur_gan,deblur_diffusion}, these models still produce hallucinations that are difficult to detect or measure. As shown in Fig.~\ref{fig:hallugen} (top), progress is limited by a \emph{circular dependency}: hallucination analysis requires labeled data, yet hallucinations are ambiguous and costly to annotate. Measuring hallucinations is non-trivial as common metrics (PSNR, SSIM, LPIPS~\cite{lpips}) favor perceptual sharpness over correctness, often scoring hallucinated outputs highly. Existing evaluations rely on downstream task performance~\cite{diffusioniqt,dynamicdps,localdiff} or manual labels~\cite{modeinterpolation,hallucination_aam}, which either conflate hallucinations with task accuracy or are restricted to simple datasets (e.g., MNIST~\cite{mnist}, Hands~\cite{hand}).

To break this deadlock, we introduce \emph{HalluGen}, a diffusion-based framework for controllable and realistic hallucination generation. Following recent work~\cite{hallucination_tomo,dynamicdps}, we categorize hallucinations as (1) \textit{intrinsic}, which break measurement consistency, and (2) \textit{extrinsic}, which remain consistent but are semantically incorrect. 

HalluGen extends diffusion-based inverse solvers with targeted gradient perturbations, enabling explicit control over hallucination \emph{type, location, and severity}. 
Controllability is crucial as hallucinations in decision-critical regions (e.g., lesions, solder joints) pose the greatest risk.
To the best of our knowledge, no prior work enables controllable, taxonomically grounded hallucination synthesis for systematic hallucination analysis in image restoration. We instantiate HalluGen with Diffusion Posterior Sampling~\cite{chung2023diffusion}, though it is compatible with other diffusion formulations.

Our contributions are threefold:
\begin{enumerate} 
    \item We propose \emph{HalluGen}, a diffusion-based framework for controllable hallucination generation in image restoration, generalizing across medical (HCP)~\cite{hcp}, industrial (MVTec AD~\cite{mvtec}), and natural (ImageNet~\cite{imagenet}) images.
    \item Using HalluGen, we construct the \emph{first large-scale hallucination dataset} for low-field MRI enhancement, comprising 4,350 predictions with patch-level annotations across 1,450 images. This dataset establishes a foundation for systematic hallucination analysis.
    \item Using the HalluGen-generated dataset, we demonstrate two key applications:
    (a) a \emph{hallucination metric benchmark}, revealing that existing metrics overlook hallucinations. We develop \emph{Semantic Hallucination Assessment via Feature Evaluation (SHAFE)}, a feature-based metric with soft-attention pooling that improves hallucination detection AUC by 0.25 and reduces false negatives by 24 percentage points; and
    (b) \emph{reference-free hallucination detectors} trained on HalluGen and evaluated on real restoration outputs without ground-truth references.
\end{enumerate}

Together, HalluGen and our dataset establish the first systematic foundation for generating and evaluating hallucinations in image restoration, advancing toward safer and more reliable deployment in safety-critical applications.

%% file: sec/2_formatting.tex
\section{Related Work}
\label{sec:formatting}

\subsection{Image Restoration}

Image restoration aims to recover clean, high-quality images from degraded measurements across tasks such as super-resolution, denoising, and deblurring.  
Modern supervised models learn powerful data-driven priors for perceptual quality~\cite{SRCNN,ESRGAN,edsr,sr_transformer,swinir}, while recent works leverage pre-trained diffusion models as generative priors for inverse problems~\cite{chung2023diffusion,chung2023solving,ddrm,zeroddrm,diffpir,stsl,daps}.  
These approaches have been extended to medical imaging, including compressed sensing and low-field MRI enhancement~\cite{iqt_pio,iqt_stochastic,uliqt,synthsr,synth_survey,dynamicdps,reconstructanything}.
However, emphasizing perceptual realism can induce hallucinations, plausible but incorrect structures, reflecting the perception–distortion trade-off~\cite{ESRGAN,perceptiontradeoff}.  
In safety-critical domains, hallucination is a known risk but remains largely unstudied, with only a few recent work~\cite{localdiff,dynamicdps,hallucination_aam,modeinterpolation} proposing mitigation strategies.

\subsection{Image Quality Metrics}

Traditional full-reference metrics~\cite{ms-ssim,vif,scc} focus on pixel-wise fidelity and global structural similarity, making them sensitive to misalignment and ineffective at capturing localized or semantically incorrect errors.
Learned perceptual metrics~\cite{lpips,dists} improve correlation with human perception by comparing deep feature representations, but they aggregate errors globally and remain insensitive to spatially sparse hallucinations.
In medical image reconstruction, recent efforts to quantify hallucinations~\cite{hallucination_metric,hallucination_tomo} are restricted to linear forward models or task-specific architectures, limiting generality.
These limitations motivate the SHAFE metric (Section~\ref{subsec:metric}), which computes local (e.g. patch) metric maps and aggregates them using soft-attention weighting, increasing sensitivity to localized hallucinations.

\subsection{Hallucination Benchmarks}

Recent benchmarks for language and vision-language models (LLMs and VLMs) evaluate semantic hallucinations such as factually incorrect or unverifiable outputs. Examples include HaluEval, HalluLens, and HallusionBench for text–image reasoning~\cite{HaluEval,hallulens,hallusionbench}, POPE for object hallucination~\cite{pope}, and VideoHallucer for temporal hallucination analysis~\cite{videohallucer}.
While some recent works~\cite{haleval,autohalluVLM,haloc} attempt to synthesize hallucinations to automate annotation in VLMs, they lack spatial control (e.g., size, location) over how hallucinations are generated.
To our knowledge, no benchmark exists for hallucinations in image restoration with spatial annotations, a gap our work addresses.

%% file: sec/3_method1.tex
\section{Method}
\label{sec:method1}

\subsection{Problem Formulation}
Image restoration aims to recover a high-quality image \(x \in \mathbb{R}^N\) from its degraded observation \(y \in \mathbb{R}^M\), formulated as an inverse problem:
\begin{equation}
    y = \mathcal{A}(x) + n,
\end{equation}
where \(\mathcal{A}(\cdot)\) denotes the forward degradation operator and \(n\) is measurement noise. As \(\mathcal{A}\) is typically non-invertible, recovery is ill-posed.

\textbf{Hallucination Taxonomy.}
We categorize hallucinations based on their relationship to the forward operator \(\mathcal{A}(\cdot)\), following the recent work on hallucination~\cite{hallucination_tomo,dynamicdps}.

\emph{Intrinsic hallucinations} violate measurement consistency, introducing features inconsistent with observed data:
\begin{equation}
\label{eq:intrinsic_definition}
    \mathcal{A}(\hat{x}) \;\neq\; \mathcal{A}(x_{\text{gt}}),
\end{equation}
where $\hat{x}$ and $x_{gt}$ denote the reconstructed image and ground truth, respectively.

\emph{Extrinsic hallucinations} preserve measurement consistency but differ in the inverse domain of \(\mathcal{A}\):
\begin{equation}
\label{eq:extrinsic_definition}
    \mathcal{A}(\hat{x}) = \mathcal{A}(x_{\text{gt}}),
    \;\; \text{but} \;\;
    \mathcal{A}^{-1}\!\big(\mathcal{A}(\hat{x})\big)
    \;\neq\;
    \mathcal{A}^{-1}\!\big(\mathcal{A}(x_{\text{gt}})\big).
\end{equation}

This distinction is crucial: intrinsic hallucinations can be detected through measurement consistency checks, while extrinsic hallucinations require ground truth or domain knowledge. In medical imaging, both pose serious risks as they may obscure clinically significant details. Understanding and systematically evaluating both categories motivates the need for controlled hallucination generation.

\subsection{Why do we need hallucination generator?}
\label{subsec:motive}

\begin{figure}[!htbp]
\centering
\includegraphics[width=\columnwidth]{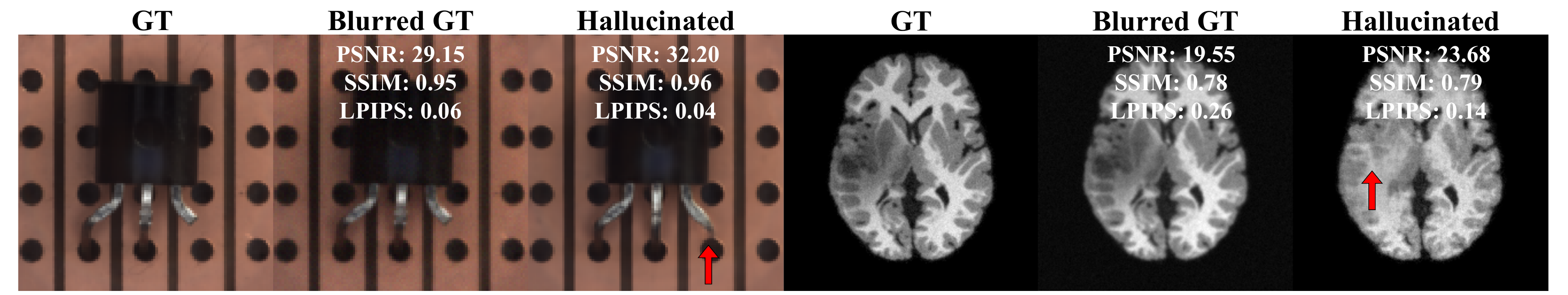}
\caption{\textbf{Existing metrics fail to penalize hallucinations.}
Across MVTec AD (left) and BraTS (right), PSNR, SSIM, and LPIPS assign higher scores to hallucinated predictions than to slightly blurred but correct images, reflecting a bias toward perceptual sharpness over correctness.}
\label{fig:motive1}
\vspace{-10pt} 
\end{figure}

\textbf{Existing metrics are insensitive to hallucinations.}
Figure~\ref{fig:motive1} shows that standard image quality metrics (PSNR, SSIM, LPIPS) assign higher scores to hallucinated predictions from pretrained conditonal diffusion models on MVTec AD~\cite{mvtec} and BraTS~\cite{brats} than to slightly blurred but semantically correct images. This exposes a critical flaw: these metrics favor perceptual sharpness over correctness of content, failing to penalize plausible yet false features.
Such bias highlights the need for hallucination-aware evaluation, which requires labeled hallucination data, a challenging circular dependency between generation and evaluation.

\textbf{Manual labeling is unreliable and uncontrolled.}
An intuitive approach to hallucination evaluation is manual annotation, but it is neither scalable nor reliable. To verify this, two domain experts annotated hallucination regions at the \emph{patch level} across 50 images, yet the inter-annotator agreement was low (Cohen’s $\kappa = 0.30$), far below the accepted threshold ($\kappa > 0.60$). This highlights the subjective and ambiguous nature of spatial hallucination identification. These limitations motivate \emph{HalluGen}, which synthesizes realistic hallucinations with known ground truth, enabling reproducible and scalable benchmarking.

\subsection{HalluGen}
\textbf{Background: Diffusion Posterior Sampling.}
We build on denoising diffusion probabilistic models (DDPMs)~\cite{ddpm} as our generative prior. Given a clean image $x_{0}$, the forward process gradually adds Gaussian noise:
\begin{equation} 
\label{eq:forward} 
q(x_{t}|x_{0}) = \mathcal{N}(x_{t}; \sqrt{\bar{\alpha}_{t}}x, (1-\bar{\alpha}_{t})I), 
\end{equation}
where $x_{t}$ denotes the noisy image at timestep $t$, $\bar{\alpha}_{t}=\prod_{i=1}^{t}(1-\beta_{i})$, $\beta_{t}$ is the noise schedule. The reverse process denoises from $t=T$ to $0$ using a learned model, where $T$ is the total number of diffusion timesteps.

To solve inverse problems, Diffusion Posterior Sampling (DPS)~\cite{chung2023diffusion} injects data consistency into the reverse process. At each step, a gradient update enforces the measurement constraint $y = \mathcal{A}(x)$:

\begin{equation} 
\label{eq:dps_update} 
x_{t-1} = \mu_{\theta}(x_{t}, t) - \lambda_{t} \nabla_{x_{t}} \|y - \mathcal{A}(\hat{x}_{0}(x_{t}))\|^{2} + \sigma_{t}\epsilon, 
\end{equation}

where $\mu_{\theta}$ is the model-predicted mean, $\hat{x}_{0}(x_{t})$ is the Tweedie estimate of the clean image, $\lambda_{t}$ controls guidance strength, and $\epsilon \sim \mathcal{N}(0, I)$. This steers sampling toward solutions consistent with the measurement $y$. 

\textbf{Intrinsic Hallucination Generation.}
HalluGen enables spatially controllable hallucination synthesis by operating at the patch level. 
It samples random binary masks \(m \in \{0,1\}^{H \times W}\) that define injection regions, varying patch sizes to create hallucinations at different spatial scales. 
To induce intrinsic hallucinations that violate measurement consistency, HalluGen applies gradient \emph{ascent} on the data-consistency term within masked regions (pushing away from consistency) and gradient \emph{descent} outside.
This process can be formulated as:
\vspace{-6pt}

\begin{align}
\label{eq:intrinsic}
x_{t-1} = \mu_{\theta}(x_{t}, t) &- \lambda_{t} \nabla_{x_{t}} \|(1-m) \odot(y - \mathcal{A}(\hat{x}_{0}(x_{t})))\|^{2} \nonumber \\
&+ \gamma_{t} \cdot \nabla_{x_{t}} \|m \odot(y - \mathcal{A}(\hat{x}_{0}(x_{t})))\|^{2},
\end{align}
where $\gamma > 0$ controls the ascent strength. This formulation creates measurement space violations $\mathcal{A}(\hat{x}) \neq \mathcal{A}(x_{\text{gt}})$ specifically within the masked patches, generating intrinsic hallucinations while preserving fidelity elsewhere.

\textbf{Extrinsic Hallucination Generation.}
Extrinsic hallucinations occur when reconstructions deviate in regions imperceptible to the forward operator while preserving measurement consistency. 
For general (nonlinear) operators \(\mathcal{A}(\cdot)\), such regions, analogous to a null space, cannot be computed explicitly. 
HalluGen therefore applies gradient \emph{ascent} in the ground-truth image space to induce semantic deviations while maintaining measurement fidelity.

However, pixel-space divergence alone yields suboptimal results, as the optimization seeks a compromise between three 
constraints: the learned data distribution \(p_{\text{data}}(x)\), measurement consistency, and ground-truth divergence. 
To address this issue, we incorporate feature-space divergence using a pretrained feature extractor \(F(\cdot)\) (e.g., DINO~\cite{dinov3}, SAM~\cite{sam}, MedSAM~\cite{medsam}):
\begin{align}
\label{eq:extrinsic}
x_{t-1} = \mu_{\theta}(x_{t}, t) &- \lambda_{t}\odot \nabla_{x_{t}} 
\|y - \mathcal{A}(\hat{x}_{0}(x_{t}))\|^{2} \nonumber \\
&+ \gamma_{1,t} \cdot \nabla_{x_{t}} \|m \odot(\hat{x}_{0}(x_{t}) 
- x_{\text{gt}})\|^{2} \nonumber \\
&+ \gamma_{2,t} \cdot \nabla_{x_{t}} \|m \odot(F(\hat{x}_{0}(x_{t})) 
- F(x_{\text{gt}}))\|^{2},
\end{align}

where $\gamma_{1,t}$ and $\gamma_{2,t}$ control pixel- and feature-space divergence strengths at time step $t$. This generates semantically meaningful hallucinations that maintain visual realism while deviating substantially in feature space.

\textbf{Hallucination Verification Module (HVM).} 
To ensure hallucinations remain visually plausible yet perceptually distinct, we measure their effect size using Cohen’s $d$ within hallucinated regions $m$ at the final diffusion step ($t=0$). Samples not meeting the taxonomy criterion are re-sampled until all patches satisfy the threshold $\tau_{\text{hvm}}$.

For \textbf{intrinsic} hallucinations, we enforce a measurement-domain violation:
\begin{equation}
d_{\text{meas}} = 
\frac{\mu_{\text{meas}}^{\text{pred}} - \mu_{\text{meas}}^{\text{gt}}}{\sigma_{\text{meas}}}
\;\ge\; \tau_{\text{hvm}}.
\label{eq:hvm_intrinsic}
\end{equation}

For \textbf{extrinsic} hallucinations, we require measurement consistency but image-domain deviation:
\begin{equation}
\begin{cases}
d_{\text{meas}} \le \tau_{\text{hvm}}, \\[6pt]
d_{\text{img}} = 
\dfrac{\mu_{\text{img}}^{\text{pred}} - \mu_{\text{img}}^{\text{gt}}}{\sigma_{\text{img}}}
\;\ge\; \tau_{\text{hvm}},
\end{cases}
\label{eq:hvm_extrinsic}
\end{equation}
where $\mu_{\text{meas}}^{\text{pred}}$, $\mu_{\text{meas}}^{\text{gt}}$, $\mu_{\text{img}}^{\text{pred}}$, and $\mu_{\text{img}}^{\text{gt}}$ are masked group means, and $\sigma_{\text{meas}}$ and $\sigma_{\text{img}}$ are the corresponding pooled standard deviations.
 Cohen’s $d$ provides a normalized effect size, making HVM domain-agnostic. $\tau_{hvm}$ controls how strictly generated samples must adhere to the taxonomy definitions.

\textbf{Manifold Regularization Effect.}
Although gradient ascent perturbs the diffusion process, generated images remain realistic due to the diffusion prior. The denoising network $\mu_{\theta}$ acts as a soft manifold prior, projecting samples toward high-likelihood regions of $p_{data}(x)$ in each time step. This preserves global coherence while introducing localized deviations, yielding plausible yet incorrect structures. 
To further enhance realism, we gradually suppress gradient ascent in the final steps, allowing the denoising prior to dominate as the process converges toward the data manifold.

\textbf{Implementation Details.}
A key design choice is \emph{where} to inject hallucinations.
Random sampling often hits background, while segmentation-based masking needs annotations.
We instead use an entropy-based strategy that selects informative regions, with patch sizes randomly drawn from 16–24 pixels to match typical hallucination extent.
For each candidate patch, we compute its Shannon entropy:
\begin{equation}
H(p) = -\sum_i p_i \log p_i,
\label{entropy}
\end{equation}
where $p_i$ is the normalized intensity histogram.
Patches failing the entropy or background-ratio thresholds (e.g., out-of-brain regions) are rejected and resampled.
This lightweight, annotation-free heuristic reliably targets semantically meaningful regions. 
More details are provided in the supplementary section.

\subsection{Dataset Construction}
\label{subsec:datasetcon}
A hallucination dataset is constructed using HalluGen on 1,450 MRI slices. 
A diffusion model is first trained on high-quality 3T brain MR images from the Human Connectome Project (HCP)~\cite{hcp}, each with resolution $256\times256$. 
To emulate low-field MRI ($<$0.36T), following DynamicDPS~\cite{dynamicdps}, a composite degradation operator is applied, 
$\mathcal{A}=\text{Blur}(\text{DS}_k(\Gamma_\gamma(\cdot)))$, with $k=4$ and $\gamma=0.7$, creating a moderately ill-posed nonlinear inverse problem suitable for controlled hallucination generation.

For each ground-truth image $x_{\text{gt}}$, three outputs are generated: 
\textbf{(1)} a non-hallucinated baseline using DPS; 
\textbf{(2)} intrinsic hallucinations; and 
\textbf{(3)} extrinsic hallucinations synthesized by HalluGen. 
Each hallucinated image contains $n\!\in\!\{1,2,3\}$ non-overlapping patches.

Outside masked regions are smoothly interpolated with the baseline at each diffusion timestep to confine evaluation to injected hallucinations. 
Each entry consists of $(x_{\text{gt}}, y, \hat{x}, m)$ and the dataset is balanced across types (1,450 intrinsic and extrinsic cases each).

%% file: sec/6_experiments.tex
\section{Experiments}
\label{sec:experiments}
\subsection{Dataset Statistics}
\label{subsec:stats}
The dataset maintains balanced coverage across hallucination characteristics. 
Intrinsic and extrinsic samples each account for 50\% of the dataset, and images with one to three hallucinated patches are evenly distributed (30--38\%). 
Spatially, hallucinations span all four image quadrants (22--30\%), ensuring diverse anatomical coverage. 
This balanced design supports fair benchmarking across diverse scenarios.

\begin{figure*}[t]
  \centering
  \includegraphics[width=0.95\linewidth]{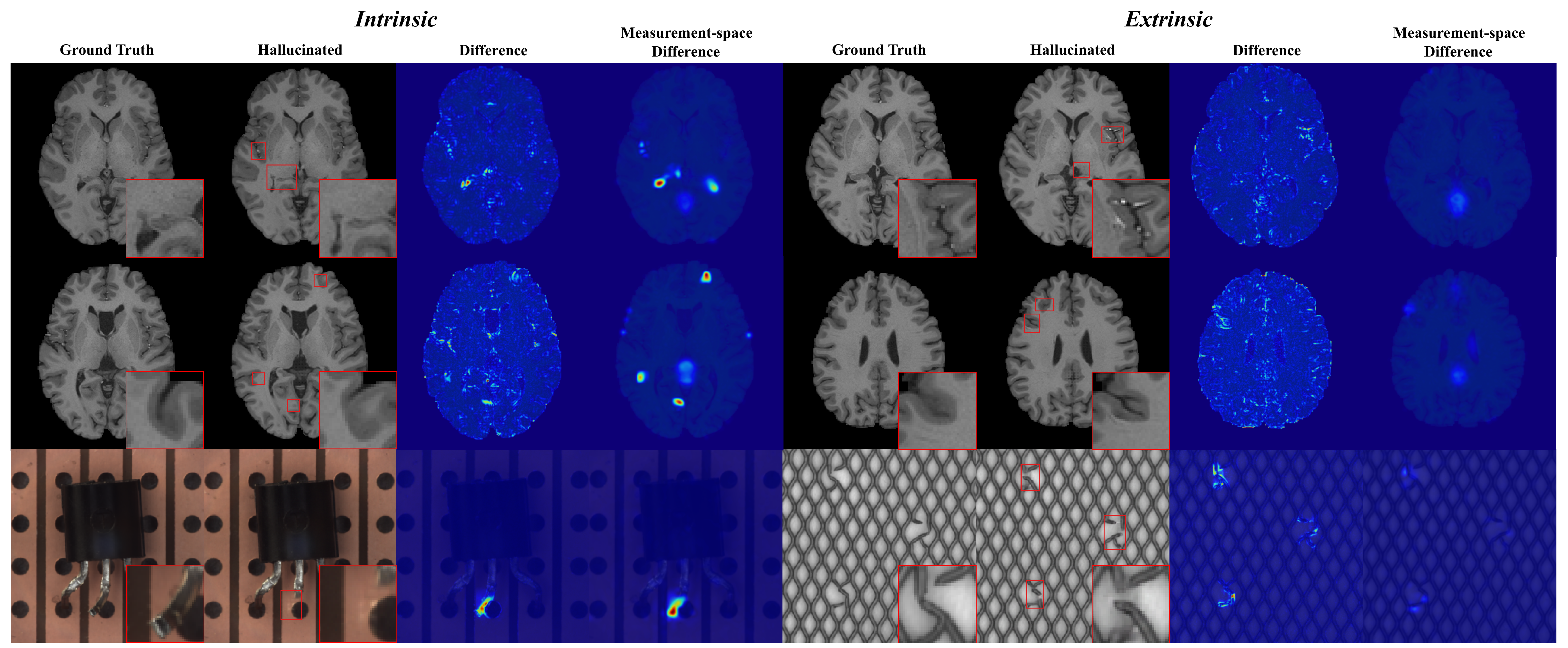}
  \caption{
  \textbf{Representative hallucinations generated by HalluGen across domains.} Top: Controlled synthesis of intrinsic and extrinsic hallucinations in our open dataset for MR images. Bottom: Cross-domain generalization to industrial imagery (MVTec AD). HalluGen produces realistic yet semantically incorrect features across domains. More visual results are in the supplementary.
  }
  \label{fig:main}
\vspace{-12pt} 
\end{figure*}

\subsection{Validation of Generated Hallucinations}
\label{subsec:validation}

Figure~\ref{fig:main} shows intrinsic and extrinsic hallucinations generated by HalluGen. The images appear realistic, altering tissue morphology such as sulci and ventricles. A key trend emerges: intrinsic hallucinations often cause morphological distortions (shape or size), whereas extrinsic ones introduce new plausible structures, leading to topological changes.

We assess perceptual realism using FID and semantic deviation via tissue segmentation IoU within hallucinated regions. Since no prior work addresses hallucination generation for image restoration, we compare HalluGen against three baselines: (1) vanilla DPS~\cite{chung2023diffusion} as a non-hallucinated reference, (2) random rotation in selected patches, and (3) measurement perturbation by adding noise in measurement space (e.g., the original input image). We additionally evaluate HalluGen with different feature extractors. Gaussian Mixture Models are used for segmenting gray matter, white matter, and CSF.

Table~\ref{tab:fid_validation} shows HalluGen achieves FID comparable to DPS, while segmentation IoU drops by \emph{50\%} within hallucinated regions, indicating strong semantic deviation. Although random rotations and measurement perturbations yield low FID, they apply only to intrinsic cases and introduce patch artifacts. HalluGen performs consistently across different feature extractors; additional visuals are provided in the supplementary. An expert study (n=2), where HalluGen outputs were mixed with vanilla DPS reconstructions, yielded only 50.5\% identification accuracy, confirming that HalluGen produces visually realistic hallucinations.

\begin{table}[h]
\centering
\footnotesize
\caption{
\textbf{Comparisons of perceptual realism and semantic deviation of HalluGen and other baselines.}
For HalluGen, FID is low, comparable to DPS, confirming realism, while segmentation IoU within hallucinated regions drops sharply across different feature extractors, verifying semantic deviation.
}

\label{tab:fid_validation}
\begin{tabular}{lccc}
\toprule
\textbf{Method} & \textbf{FID} $\downarrow$ & \textbf{IoU} $\downarrow$ & \textbf{Tax. Comp.} \\
\midrule
DPS~\cite{chung2023diffusion} & 0.32 & 0.861$\pm$0.09 & - \\
Random Rotation & 0.32 & 0.393$\pm$0.12 & Intr. only \\
Meas. Perturb. & 0.48 & 0.721$\pm$0.10 & Intr. only \\
\textbf{HalluGen + MedSAM~\cite{medsam}} & 0.41 & 0.363$\pm$0.15 & Both \\
\textbf{HalluGen + SAM~\cite{sam}} & 0.36 & 0.367$\pm$0.15 & Both \\
\textbf{HalluGen + DINOv3~\cite{dinov3}} & 0.43 & 0.362$\pm$0.04  & Both \\
\bottomrule
\end{tabular}
\vspace{-8pt} 
\end{table}

We validate HalluGen’s taxonomy compliance in Tab.~\ref{tab:taxonomy} by measuring measurement- and image-space deviations within masked regions. Intrinsic hallucinations exhibit measurement loss roughly \emph{7 times} higher than DPS, confirming violation of measurement consistency. Extrinsic hallucinations retain low measurement loss but diverge in image space, consistent with the formal taxonomy (Eq.~\ref{eq:intrinsic_definition}--\ref{eq:extrinsic_definition}). Fig.~\ref{fig:main} further visualizes measurement-space difference maps, with extrinsic cases showing much lower errors than intrinsic ones within hallucinated regions.

\begin{table}[!hbpt]
\centering
\footnotesize
\caption{
\textbf{Hallucination taxonomy compliance of HalluGen using Mean Squared Error within masked region.} Intrinsic sustains high measurement loss while extrinsic maintains low measurement loss despite semantic errors in image space. This validation confirms HalluGen faithfully generates both taxonomy types.
}
\label{tab:taxonomy}
\begin{tabular}{lcc}
\toprule
\textbf{Predicted Region} & \textbf{Meas. Loss} & \textbf{Image Loss} \\
\midrule
DPS & 0.006 & 0.022 \\
Intrinsic & 0.039 & 0.058 \\
Extrinsic & 0.003 & 0.041 \\
\bottomrule
\end{tabular}
\vspace{-12pt} 
\end{table}

\subsection{Controllability of HalluGen}
\label{subsec:control}
HalluGen enables fine-grained control over hallucination characteristics. We validate controllability along three dimensions: gradient strength, patch count, and patch size. We focus on intrinsic hallucinations, as extrinsic generation involves balancing multiple loss terms that complicate the isolation of individual control factors.

\begin{figure*}[!htbp]
\centering
\includegraphics[width=0.33\textwidth]{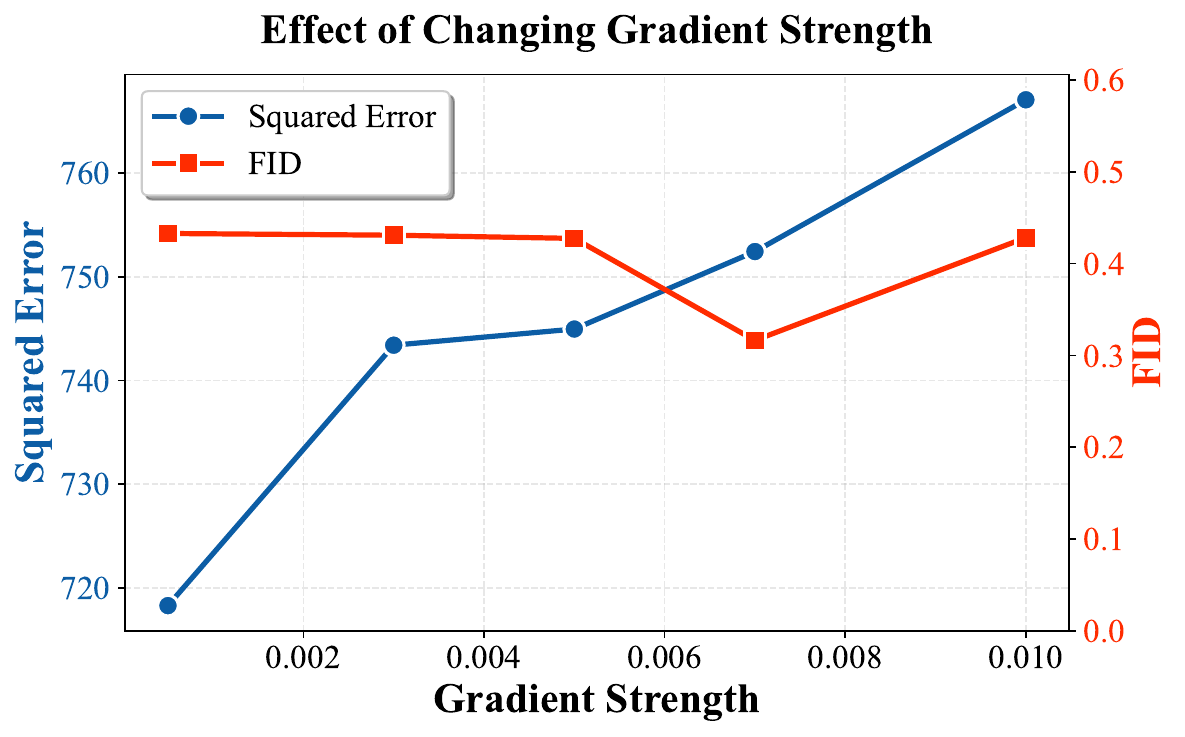}
\hfill
\includegraphics[width=0.33\textwidth]{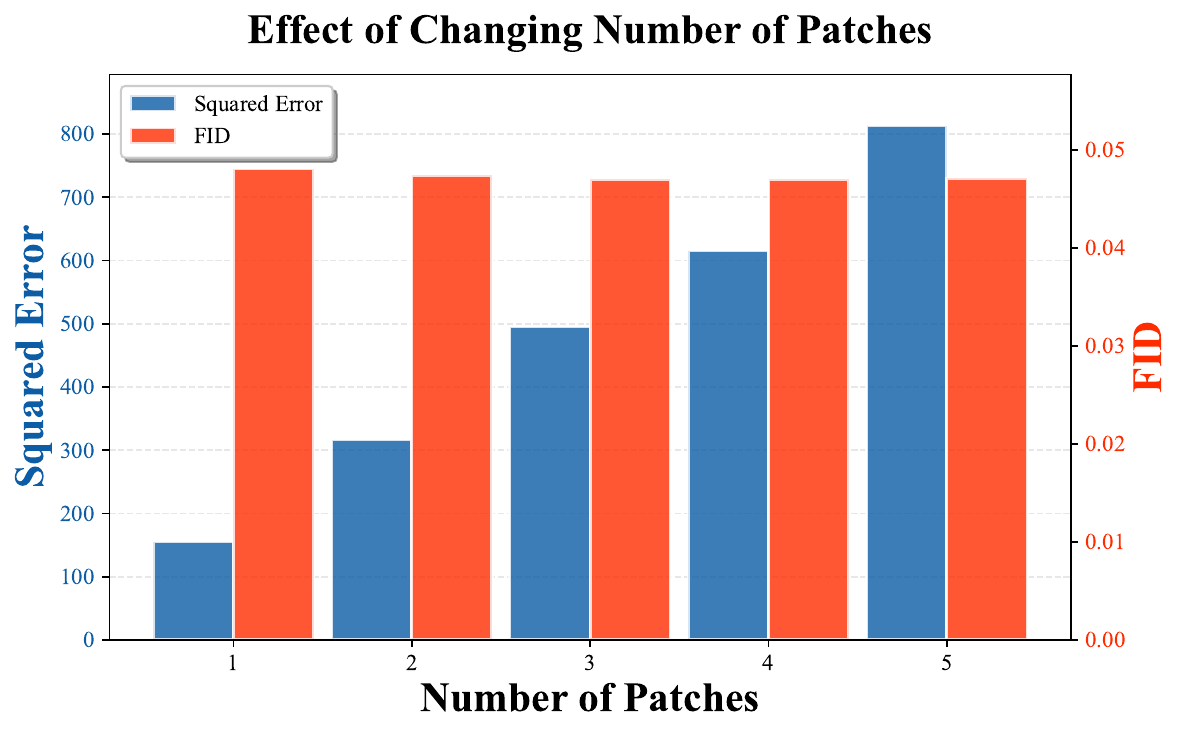}
\hfill
\includegraphics[width=0.33\textwidth]{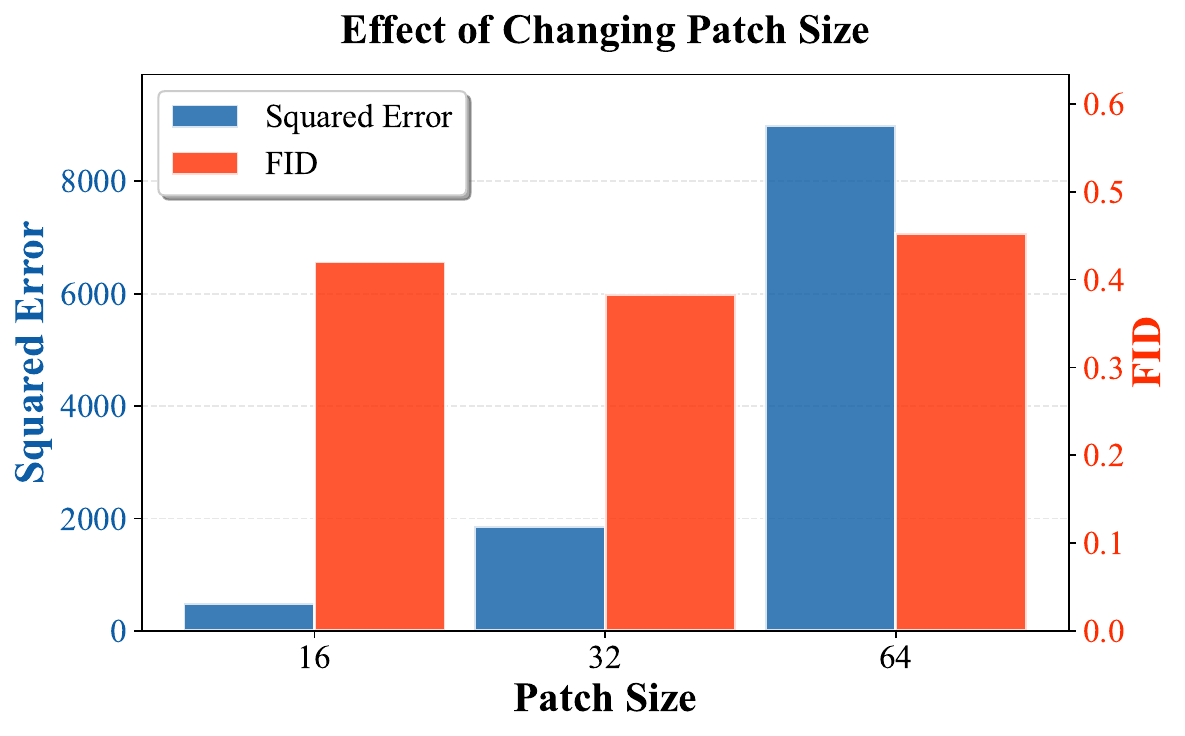}
\caption{
\textbf{Controllability of HalluGen.}
Left: Severity increases with gradient strength $\gamma$.
Middle: Severity scales linearly with number of patches while FID stays low.
Right: Stable realism across patch sizes (16×16 – 64×64).
HalluGen provides fine-grained control over severity, spatial extent, and granularity while preserving realism.
}
\label{fig:controllability}
\vspace{-12pt} 
\end{figure*}

\textbf{Severity control via gradient strength.} Figure~\ref{fig:controllability} (left) shows that hallucination severity increases with gradient ascent strength, $\gamma$. The squared error in hallucinated regions increases from 718 ($\gamma=0.0005$) to 767 ($\gamma=0.01$), showing that stronger gradients induce more severe, measurement-inconsistent hallucinations. FID remains stable, indicating preserved perceptual realism.

\textbf{Spatial scalability via number of patches.} Figure~\ref{fig:controllability} (middle) validates that HalluGen scales with the number of patches. As the number increases from 1 to 5, total squared error grows approximately linearly (150 to 810). FID remains consistently low, confirming that distributed hallucinations maintain realistic appearances.

\textbf{Patch size flexibility.} Figure~\ref{fig:controllability} (right) shows controllability across different spatial granularities. Squared error scales with patch area (16×16: 488, 64×64: 9000), while FID remains stable, demonstrating consistent realism across fine-grained and coarse spatial scales.

These results show that HalluGen enables precise control over hallucination severity, extent, and scale while preserving perceptual realism through manifold regularization.

\subsection{Cross-Domain and Cross-Task Generalization}
\label{subsec:domain}
\begin{figure}[!hbtp]
\centering
\includegraphics[width=\columnwidth]{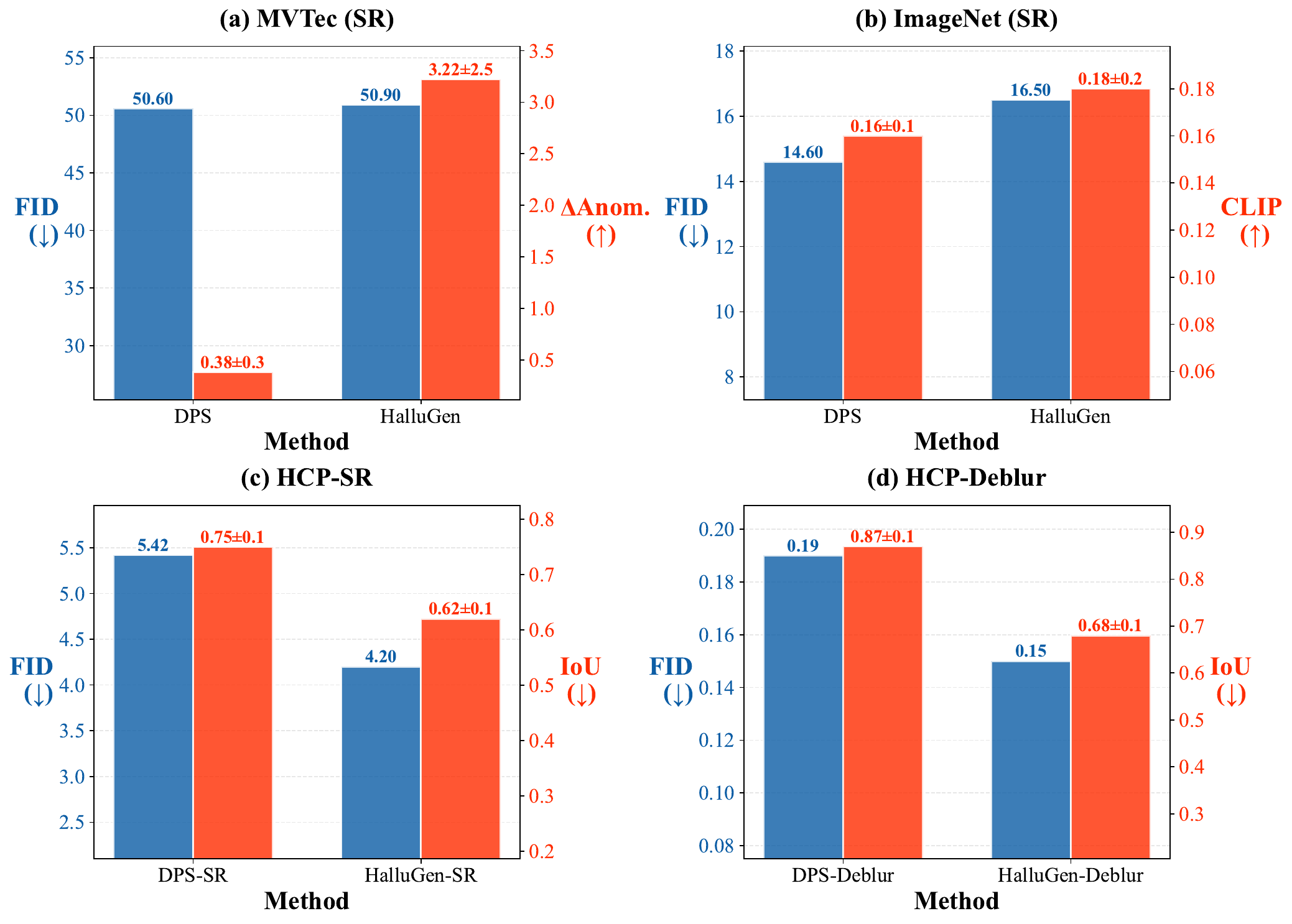}
\caption{
\textbf{Cross-domain and cross-task generalization of HalluGen.}
Low FID and high semantic deviations confirm HalluGen’s generalizability. 
For SR, we use $\times6$; for deblur, $\sigma\!=\!3.0$.
}
\label{fig:cross_generalization}
\vspace{-16pt} 
\end{figure}

We evaluate HalluGen’s generalizability by sampling 200 images per domain-task pair (1600 total) for both intrinsic and extrinsic, summarized in Fig.~\ref{fig:cross_generalization}. 
For MVTec AD~\cite{mvtec}, semantic deviation is measured by the change in anomaly scores between ground-truth and generated images using the pre-trained PatchCore~\cite{patchcore}. 
On ImageNet~\cite{imagenet}, it is quantified by the CLIP score, the distance between CLIP embeddings of the ground-truth and generated images.

\begin{figure}[!hbtp]
\centering
\includegraphics[width=0.95\columnwidth]{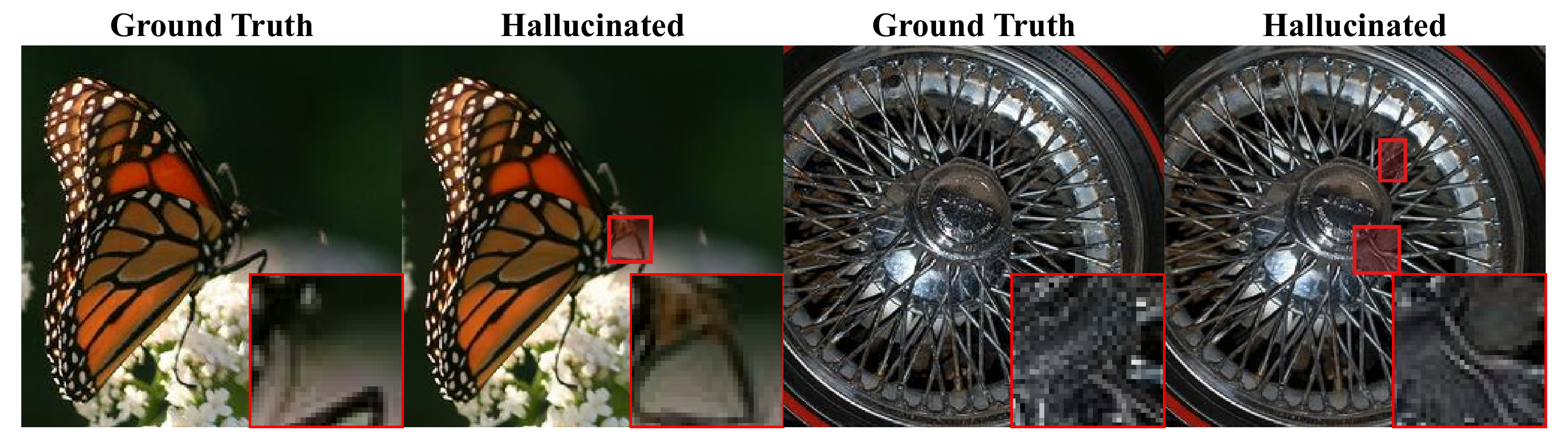}
\caption{
\textbf{Visual results of HalluGen on ImageNet.}
HalluGen generalizes beyond safety-critical domains and can generate realistic hallucinated features as highlighted in red boxes.
}
\label{fig:imagenet}
\vspace{-12pt} 
\end{figure}

Fig.~\ref{fig:main} shows HalluGen generating hallucinated features on MVTec AD, such as missing transistor legs and distorted grid patterns. 
It maintains low FID with large change in anomaly score (Fig.~\ref{fig:cross_generalization}), indicating perceptually plausible but semantically divergent outputs. 
HalluGen also generalizes to natural images, generating realistic distortions such as extra insect legs, warped tires with an increase in CLIP score compared to DPS (Figs.~\ref{fig:imagenet}, \ref{fig:cross_generalization}).

Additional experiments on HCP~\cite{hcp} confirm robustness across standard restoration tasks, including super-resolution and deblurring, showing that HalluGen generalizes across both domains and restoration types.

\subsection{Case Study I: Hallucination Metric Benchmark}
\label{subsec:metric}
The evaluation of hallucination has been hindered by the lack of labeled data. 
Leveraging HalluGen-generated dataset, we establish the first benchmark for evaluating hallucination sensitivity of full-reference image quality metrics. 
We assess representative metrics by discriminability (Cohen’s~$d$), detectability (AUC-ROC), sharpness bias curve (SBC), and correlation with hallucination severity (Spearman’s~$\rho$).
SBC is computed as the area under the win-rate curve obtained by progressively low-pass filtering non-hallucinated images. Severity is defined as the sum of squared error within hallucinated regions scaled by the number of hallucinated patches, to penalize images with more frequent hallucination occurrences.

We also develop a hallucination-aware baseline, \emph{Semantic Hallucination Assessment via Feature Evaluation (SHAFE)}, to address the limitation of conventional metrics on hallucinations and evaluate it alongside existing metrics.

\textbf{Semantic Hallucination Assessment via Feature Evaluation (SHAFE).}
Most full-reference metrics use uniform averaging over errors, which suppresses hallucinations that are \emph{localized and sparse}. SHAFE applies soft attention–based aggregation~\cite{attentionpooling}, emphasizing rare but structurally inconsistent regions while preserving global context.
Given a prediction $\hat{x}$ and ground truth $x_\text{gt}$, SHAFE is computed as follows:

• \textbf{Patch-level feature discrepancy.} A low-pass filter is applied to both inputs to suppress high-frequency noise and emphasize structural content. Each image is then fed into a pretrained vision encoder to extract semantic features, and cosine distance $\delta_{cos}$ is computed for each patch. Shallow multi-layer features are extracted to retain spatial richness and reduce dataset-specific bias of the encoder.

\begin{figure}[!htbp]
\centering
\includegraphics[width=0.48\columnwidth]{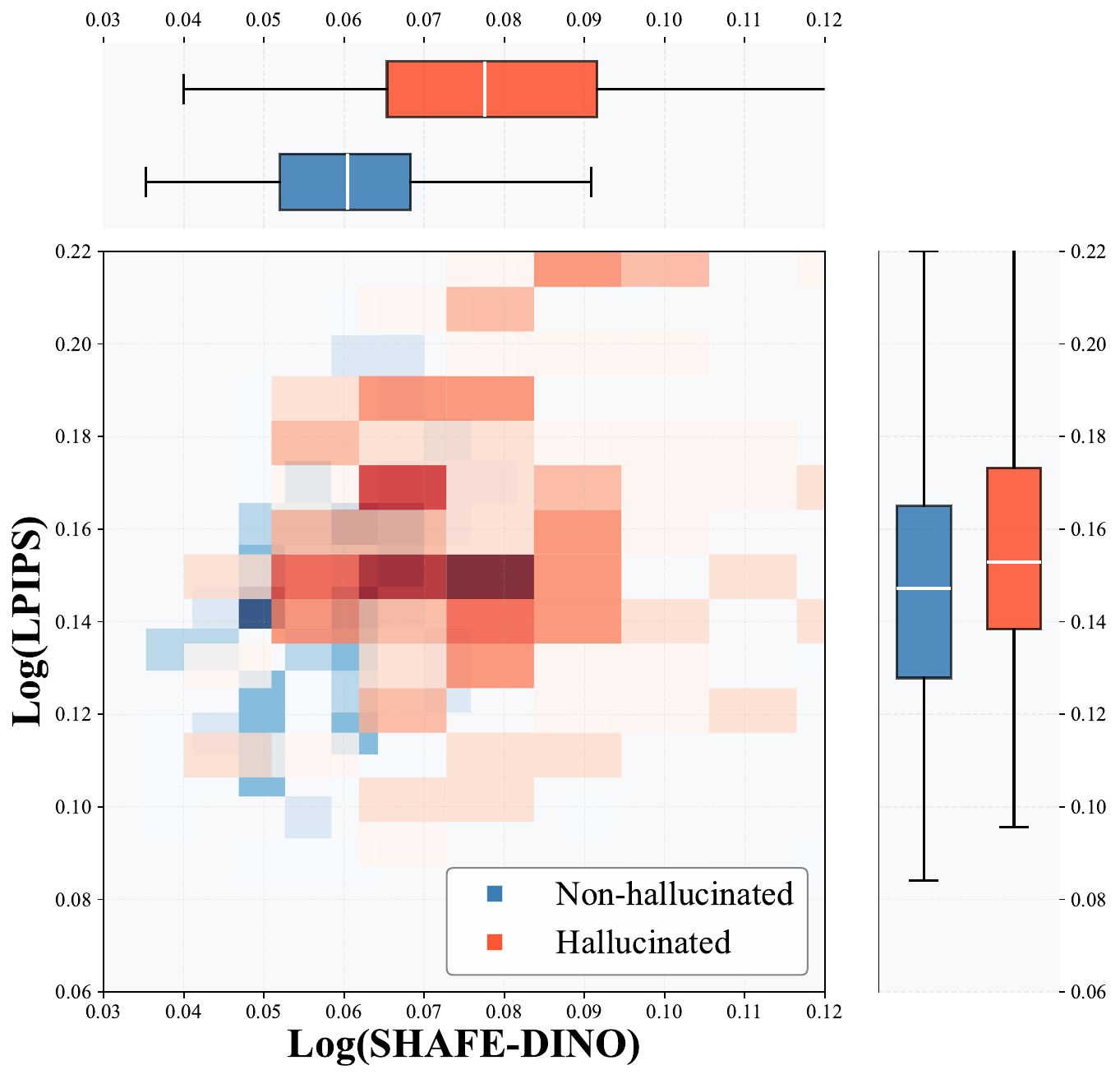}
\hfill
\includegraphics[width=0.44\columnwidth]{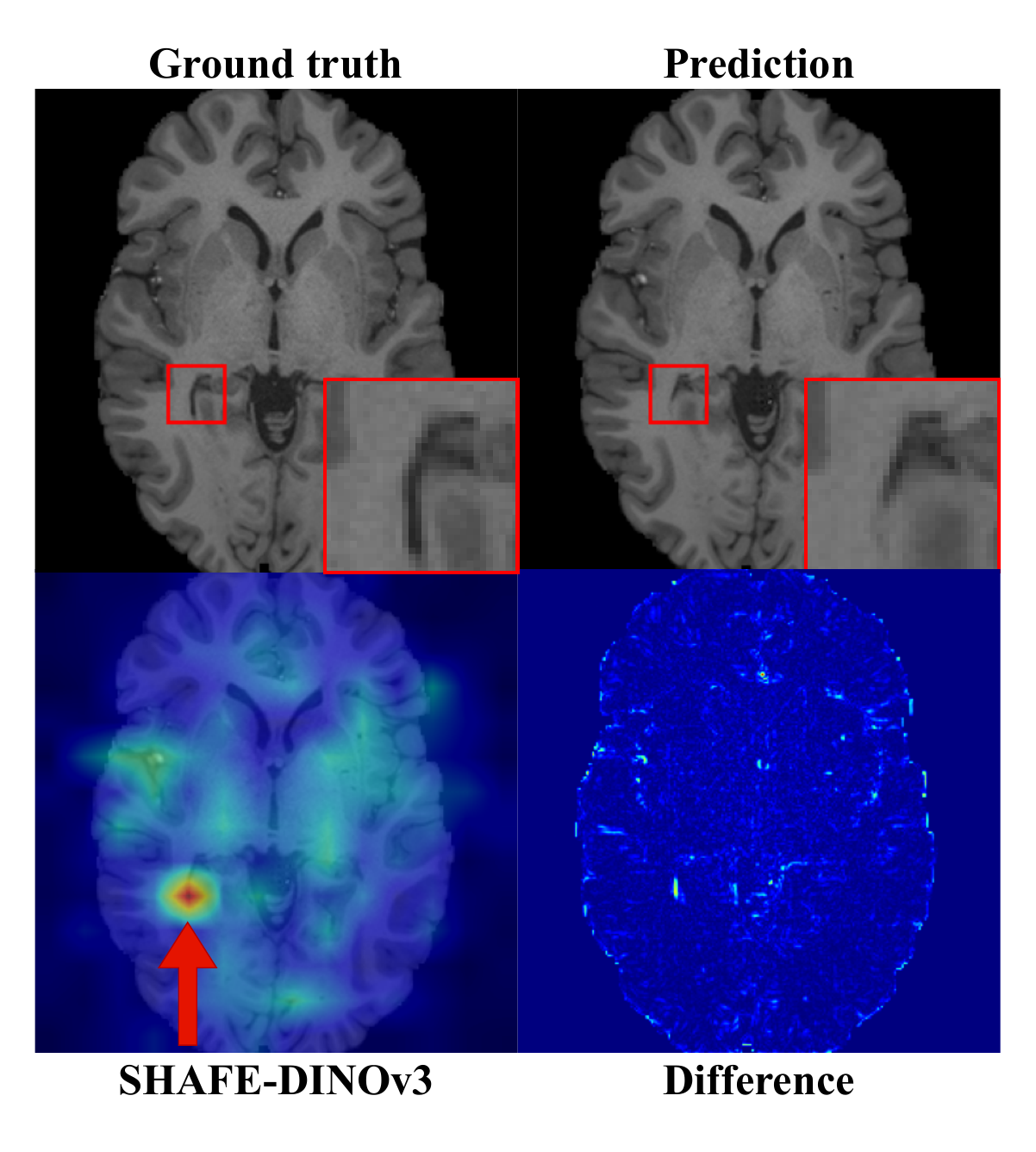}
\caption{
\textbf{Bivariate distributions of SHAFE-DINOv3 vs. LPIPS (left) and Spatial localization ability of SHAFE on \emph{real prediction} (right).} Hallucinated examples shift toward higher SHAFE values, indicating improved separability compared to LPIPS.
}
\label{fig:real_auc}
\vspace{-6pt} 
\end{figure}

• \textbf{Softmax-based aggregation.}
These patch scores are then aggregated using temperature-controlled softmax:
\begin{equation}
\label{equ:aggregation}
\text{SHAFE} = \sum_i w_i \, \delta_{cos,i}, \quad
w_i = \frac{\exp(\delta_{cos,i} / \tau)}{\sum_j \exp(\delta_{cos,j} / \tau)}.
\end{equation}
Smaller $\tau$ assigns exponentially greater weights, $w_{i}$, to local and severe errors. Its patch-based design enables hallucination localization, with Fig.~\ref{fig:real_auc} showing SHAFE heatmaps accurately highlighting real failure regions.

SHAFE operates in feature space, but its formulation is general and can be applied to other metrics. Moreover, its differentiability allows integration into supervised training pipelines, opening avenues for future extensions toward hallucination-aware model optimization. See the supplementary material for more details.

\textbf{Perceptual metrics are insensitive to hallucinations;} Tab.~\ref{tab:metric_benchmark} summarizes the performance of the metrics on hallucination. LPIPS and DISTS perform marginally above chance in hallucination detection, showing limited sensitivity. Their \emph{low} Cohen’s $|d|$ and \emph{high} per-sample variability reveal inconsistent responses across hallucinations, making them unreliable indicators. Because they uniformly average spatial errors and rely on ImageNet-trained features, these metrics favor perceptual sharpness over semantic correctness and often overlook structurally inconsistent regions.

\textbf{Multi-scale improves sensitivity.} MS-SSIM substantially improves hallucination sensitivity over SSIM, benefiting from its multi-scale formulation that captures hallucinations of varying sizes. This demonstrates the value of incorporating multi-scale information for more reliable image evaluation. However, it still underperforms SHAFE due to its uniform aggregation strategy.

\textbf{Pixel metrics show type-dependent performance.} Pixel-based metrics achieve moderate detection on intrinsic hallucinations that violate measurement consistency and produce large pixel-space deviations. However, their performance degrades substantially on extrinsic cases that remain measurement-consistent despite semantic incorrectness, limiting their utility for comprehensive evaluation.

\begin{table*}[!htbp]
\centering
\footnotesize
\caption{
\textbf{Comprehensive hallucination benchmark across metrics.}
Comparison of pixel- and feature-based metrics on effect size (discrimination), AUC (detection), Sharpenss Bias Curve (sharpness bias), and severity correlation (spearman rank correlation). The rightmost column shows AUC on raw predictions with manually labeled (binary) hallucinations. Baseline metrics show weak hallucination sensitivity, whereas SHAFE improve detection and interpretability. FPR/FNR are calculated using the optimal threshold from AUC curve. 
}

\label{tab:metric_benchmark}
\begin{tabular}{l|c|cc|cc|cc|cc: c|c}
\toprule
\multirow{2}{*}{\textbf{Metric}} & \multirow{2}{*}{\textbf{Type}} 
& \multicolumn{2}{c|}{\textbf{Effect Size (Cohen's $|d|$) $\uparrow$}} 
& \multicolumn{2}{c|}{\textbf{AUC-ROC $\uparrow$}} 
& \multicolumn{2}{c|}{\textbf{SBC $\uparrow$}} 
& \multicolumn{2}{c:}{\textbf{Severity ($\rho$) $\uparrow$}} 
& \textbf{FNR/FPR}
& \textbf{Real} \\
& & Intr. & Extr. & Intr. & Extr. & Intr. & Extr. & Intr. & Extr. & \textbf{(\%)$\downarrow$} & \textbf{AUC$\uparrow$}\\
\midrule
PSNR & Pixel & 0.33$\pm$0.95 & 0.02$\pm$0.99 & 0.59 & 0.51 & 0.76 & 0.49 & 0.37 & 0.32 & 62 / 28 & 0.58 \\
SSIM & Pixel & 0.09$\pm$0.99 & 0.33$\pm$1.03 & 0.52 & 0.40 & 0.22 & 0.08 & 0.11 & -0.17 & 99 / 0 & 0.53 \\
MS-SSIM~\cite{ms-ssim} & Pixel & 0.69$\pm$1.00 & 0.16$\pm$0.99 & 0.79 & 0.21 & \textbf{0.84} & 0.27 & 0.43 & 0.08 & 36 / 57 & 0.57 \\
SCC~\cite{scc} & Pixel & 0.06$\pm$1.03 & 0.16$\pm$1.01 & 0.48 & 0.46 & 0.23 & 0.12 & 0.05 & 0.02 & 80 / 18 & 0.47 \\
LPIPS~\cite{lpips} & Feature & 0.03$\pm$1.00 & 0.01$\pm$0.99 & 0.51 & 0.50 & 0.14 & 0.12 & 0.11 & 0.14 & 26 / 70 & 0.42 \\
DISTS~\cite{dists} & Feature & 0.06$\pm$0.99 & 0.10$\pm$0.99 & 0.52 & 0.54 & 0.27 & 0.27 & 0.19 & 0.12 & 31 / 60 & 0.44 \\
VIF~\cite{vif} & Pixel & 0.29$\pm$1.00 & 0.00$\pm$0.99 & 0.58 & 0.50 & 0.28 & 0.11 & 0.27 & 0.17 & 42 / 48 & 0.47 \\
\midrule
\textbf{SHAFE-ResNet50~\cite{resnet50aa}} & Feature & \textbf{1.18$\pm$1.12} & 0.84$\pm$1.28 & \textbf{0.82} & 0.72 & 0.79 & \textbf{0.71} & \textbf{0.51} & 0.57 & 27 / 25 & 0.58 \\
\textbf{SHAFE-MedSAM~\cite{medsam}} & Feature & 1.13$\pm$1.26 & \textbf{0.93$\pm$1.34} & 0.81 & \textbf{0.75} & 0.77 & 0.71 & 0.47 & \textbf{0.60} & \textbf{32 / 14} & 0.70 \\
\textbf{SHAFE-DINOv3~\cite{dinov3}} & Feature & 1.12$\pm$1.17 & 0.92$\pm$1.30 & 0.76 & 0.74 & 0.71 & 0.67 & \textbf{0.51} & 0.58 & 30 / 26 & \textbf{0.79} \\
\bottomrule
\end{tabular}
\vspace{-10pt} 
\end{table*}

\textbf{SHAFE achieves superior performance.} 
SHAFE variants outperform baselines, with \emph{all feature backbones} improving all benchmark metrics by 30\% over the best baselines. 
Especially, SHAFE-ResNet50 outperforms LPIPS and DISTS, despite sharing the same pre-training domain, highlighting the effectiveness of our weighted aggregation to emphasize local failures. 
Moreover, all SHAFE variants exhibit low sharpness bias, confirming that the combination of low-pass filtering and our multi-layer feature extraction strategy reduces the reliance on high-frequency artifacts.

We evaluated metrics on real restoration outputs by annotating 300 predictions from SwinIR~\cite{swinir}, EDSR~\cite{edsr}, and ESRGAN~\cite{ESRGAN}. SHAFE shows much stronger discrimination, improving AUC by 0.2 over traditional metrics (Table~\ref{tab:metric_benchmark}), aligning more closely with human judgment. Figure~\ref{fig:real_auc} shows that SHAFE provides distinctly better separation between images with and without hallucinations than LPIPS, demonstrating its robustness in real cases.

\begin{figure}[!hbtp]
\centering
\includegraphics[width=\columnwidth]{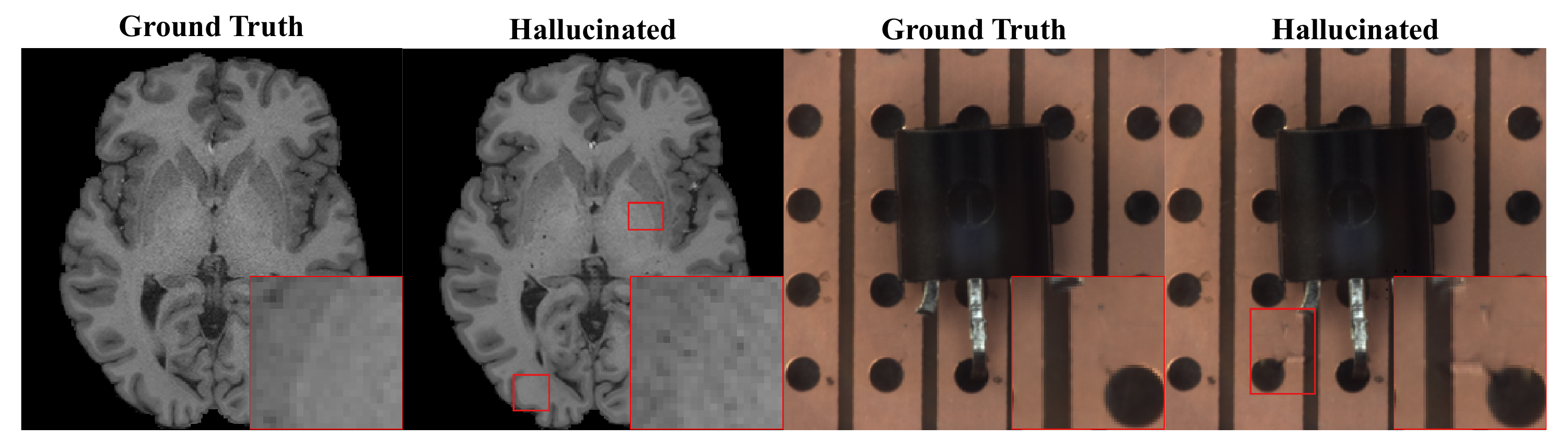}
\caption{
\textbf{Failure case of HalluGen in homogeneous regions.}
In smooth regions (e.g., brain putamen or background), diffusion prior dominates, leading to visually weak hallucinations. 
}
\label{fig:failure}
\vspace{-12pt} 
\end{figure}

\subsection{Case Study II: Hallucination Detection}
\label{subsec:detection}
HalluGen's utility can be extended to developing data-driven hallucination detector.
We train a \emph{reference-free hallucination detector} using a CNN that takes the prediction $\hat{\mathbf{x}}$ and measurement $\mathbf{y}$ as input and predicts whether hallucinations are present, without requiring ground-truth images. The model is trained on the full HalluGen dataset (4,350 images) and evaluated on: (1) a held-out HalluGen split (702 images; evenly balanced across intrinsic, extrinsic, and non-hallucinated cases), and (2) same real restoration outputs used in the benchmark in Tab.~\ref{tab:metric_benchmark}. 

As shown in Table~\ref{tab:detection}, the detector achieves strong performance on synthetic data and generalizes to real predictions, despite its simple architecture. Detection is notably easier for intrinsic hallucinations than extrinsic by AUC $\approx$ 0.14. This is because intrinsic hallucinations break measurement consistency, whereas extrinsic ones remain consistent with $y$ and are therefore harder to identify. Overall, these results show that HalluGen \emph{transfers to real model failures} and enables comprehensive, systematic hallucination evaluation that was previously infeasible.

\begin{table}[!htbp]
\vspace{-8pt} 
\centering
\footnotesize
\caption{\textbf{Reference-free hallucination detection.} CNN trained only on HalluGen; thresholds selected on validation.}
\label{tab:detection}
\begin{tabular}{lcc}
\toprule
\textbf{Test Set} & \textbf{AUC} $\uparrow$ & \textbf{F1} $\uparrow$ \\
\midrule
HalluGen (Intrinsic) & \textbf{0.91} & \textbf{0.95} \\
HalluGen (Extrinsic) & 0.77 & 0.76 \\
Real Predictions     & 0.73 & 0.63 \\
\bottomrule
\end{tabular}
\vspace{-12pt} 
\end{table}

%% file: sec/8_conclusion.tex
\section{Conclusion}
\label{sec:conclusion}
We present \emph{HalluGen}, a principled framework for controllable hallucination synthesis, and construct the first hallucination dataset for image restoration with patch-level annotations. 
We demonstrate HalluGen's utility in two key applications: 
(1) establishing a hallucination-metric benchmark that reveals the failure of existing image quality metrics and developing \emph{SHAFE}, which improves hallucination sensitivity via soft-attention pooling over feature errors; and 
(2) training reference-free hallucination detectors that generalize to real restoration outputs. 
These results demonstrate that \emph{HalluGen transfers to real model failures, enabling systematic study of hallucinations in real-world settings.}
HalluGen has several limitations, including difficulty inducing hallucinations in homogeneous regions due to strong diffusion priors (Fig.~\ref{fig:failure}) and lack of explicit semantic control. 
The current dataset is restricted to MR images, and extending HalluGen to other safety-critical domains remains an important direction. 
Nevertheless, HalluGen establishes the first systematic foundation for generating, detecting, and evaluating hallucinations in image restoration. 
We release our code, dataset, and metrics to support further research and safer deployment in safety-critical domains.

%% file: sec/X_suppl.tex
\clearpage
\setcounter{page}{1}
\maketitlesupplementary

\section*{Table of Contents}

\begin{itemize}[leftmargin=1.2em]

\item \textbf{6. HalluGen}
    \begin{itemize}[leftmargin=1.2em]
        \item Sec.~\ref{subsec:hallugen_details} Implementation Details
        \item Sec.~\ref{subsec:hallugen_ablations} Ablations
        \item Sec.~\ref{subsec:hallugen_visual} Qualitative Results
    \end{itemize}

\item \textbf{7. Dataset}
    \begin{itemize}[leftmargin=1.2em]
        \item Sec.~\ref{subsec:dataset_instruction} Instructions for Dataset
        \item Sec.~\ref{subsec:dataset_computational} Computational Cost
        \item Sec.~\ref{subsec:dataset_limitaiton} Limitations
        \item Sec.~\ref{subsec:dataset_license} License and Ethics Statement
    \end{itemize}

\item \textbf{8. SHAFE}
    \begin{itemize}[leftmargin=1.2em]
        \item Sec.~\ref{subsec:shafe_details} Implementation Details
        \item Sec.~\ref{subsec:shafe_ablation} Ablations
        \item Sec.~\ref{subsec:shafe_localization} Localization Results
        \item Sec.~\ref{subsec:shafe_failure} Limitations of SHAFE 
        \item Sec.~\ref{subsec:shafe_computational} Inference Speed
    \end{itemize}

\item \textbf{\ref{sec:hallu_detector}. Hallucination Detectors}

\item \textbf{Figures}
    \begin{itemize}[leftmargin=1.2em]
        \item Fig.~\ref{fig:tau_effect}: Effect of temperature $\tau$ in softmax aggregation for SHAFE (AUC vs. $\tau$)
        \item Fig.~\ref{fig:localization}: Visual results of SHAFE heatmap on both HalluGen and real restoration outputs
        \item Fig.~\ref{fig:failure}: Visual examples of SHAFE limitations
        \item Fig.~\ref{fig:inf_time}: Inference time comparison for baseline metrics vs. SHAFE-ResNet50
        \item Fig.~\ref{fig:detector}: Network archtiecture of reference-free hallucination detector
        \item Fig.~\ref{fig:hcp_total}: Visual results of HalluGen on brain MRI
        \item Fig.~\ref{fig:mvtec_total}: Visual results of HalluGen on MVTec AD
        \item Fig.~\ref{fig:imagenet_total}: Visual results of HalluGen on ImageNet
    \end{itemize}

\item \textbf{Tables}
    \begin{itemize}[leftmargin=1.2em]
        \item Tab.~\ref{tab:hvm}: HalluGen ablations using mean squared error within masked region
        \item Tab.~\ref{tab:shafe}: SHAFE ablations using AUC on hallucination detection
        \item Tab.~\ref{tab:feature}: Impact of feature-layer selection on SHAFE hallucination detection
    \end{itemize}

\end{itemize}

\section{HalluGen}
\subsection{Implementation Details}
\label{subsec:hallugen_details}
We provide additional implementation details for HalluGen. Intrinsic and extrinsic generation procedures are summarized in Algo.~\ref{alg:hallugen_intrinsic} and Algo.~\ref{alg:hallugen_extrinsic}. We first obtain a non-hallucinated baseline $x_{\text{base}}$ using DPS~\cite{chung2023diffusion}, then sample candidate patches for hallucination injection. Patches failing entropy or background thresholds are rejected and resampled. HalluGen is initialized either from Gaussian noise (when $t_{\text{skip}}=\text{None}$) or from a partially noised version of $x_{\text{base}}$ at timestep $t_{\text{skip}}$.

At each reverse step, we compute the Tweedie estimate $\hat{x}_{0}(x_{t})$. When interpolation is enabled, $\hat{x}_{0}(x_{t})$ is blended with $x_{\text{base}}$ outside the mask to preserve fidelity in non-masked regions. Intrinsic generation modifies only the measurement-consistency term, whereas extrinsic additionally applies a feature-space loss.

To maintain realism and avoid patch artifacts, all ascent weights are set to zero after timestep $t_{\text{stop}}$, following LocalDiffusion~\cite{localdiff}, allowing the diffusion prior to refine boundaries in later steps. Finally, the Hallucination Verification Module (HVM) checks whether masked deviations satisfy the taxonomy criteria (Eqs.~\ref{eq:intrinsic_definition}, \ref{eq:extrinsic_definition}) via Cohen’s $d$; samples failing the threshold $\tau_{\text{hvm}}$ are rejected and regenerated.

\noindent\textbf{Hyperparameters.} We use the following default settings:
\begin{itemize}[leftmargin=1.5em]
    \item Diffusion solver: DDPM~\cite{ddpm}
    \item Diffusion parameterization: $\varepsilon_{\theta}$
    \item Diffusion steps: $T = 1000$
    \item Skip timestep: $t_{\text{skip}} = 200$
    \item Measurement-consistency weight: $\lambda_t = 1.0$
    \item Intrinsic ascent weight: $\gamma_t = 0.002$
    \item Extrinsic ascent weights: $\gamma_{1,t}, \gamma_{2,t} = 0.007$
    \item Feature-loss encoder: MedSAM~\cite{medsam}
    \item Entropy / background thresholds: $1.4 / 0.05$
    \item HVM threshold: $\tau_{\text{hvm}} = 0.1$
    \item Patch size: 16--24
    \item Number of patches: 1--3
    \item Ascent cut-off timestep: $t_{\text{stop}} = 10$
\end{itemize}

\begin{algorithm}[t]
\caption{HalluGen -- Intrinsic Hallucination}
\label{alg:hallugen_intrinsic}
\begin{algorithmic}[1]
\Require Ground truth $x_{\text{gt}}$, measurement $y$, forward operator $\mathcal{A}(\cdot)$
\Require Diffusion model $\mu_{\theta}$, step size $\{\lambda_t\}$, $\{\gamma_t\}$
\Require Non-hallucinated baseline $x_{\text{base}}$
\Require Optional skip timestep $t_{\text{skip}}$ (or None), ascent cut-off timestep $t_{\text{stop}}$
\Require Interpolation flag $\text{interp} \in \{\text{true},\text{false}\}$, interpolation schedule $w_t$
\State Sample hallucination mask $m \in \{0,1\}^{H \times W}$ by entropy-based patch selection
\If{$t_{\text{skip}}$ is None}
    \State $t_{\text{start}} \gets T$
    \State Sample $x_T \sim \mathcal{N}(0, I)$
\Else
    \State $t_{\text{start}} \gets t_{\text{skip}}$
    \State Sample $x_{t_{\text{start}}} \sim q(x_{t_{\text{start}}} \mid x_{\text{base}})$ \Comment{forward process $x_{\text{base}}$}
\EndIf
\For{$t = t_{\text{start}}, t_{\text{start}}-1, \ldots, 1$}
    \State Compute Tweedie estimate $\hat{x}_0(x_t)$
    \If{$\text{interp} = \text{true}$}
        \State $\hat{x}_0(x_t) \gets (1 - m)\odot\big[w_t\,\hat{x}_0(x_t) + (1 - w_t)\,x_{\text{base}}\big] + m \odot \hat{x}_0(x_t)$
    \EndIf
    \State $x_{t-1} \gets \mu_{\theta}(x_t, t) - \lambda_t \nabla_{x_t} \big\|(1-m)\odot \big(y - \mathcal{A}(\hat{x}_0)\big)\big\|_2^2$
    \If{$t > t_{\text{stop}}$}
        \State $\gamma_t^{\text{eff}} \gets \gamma_t$
    \Else
        \State $\gamma_t^{\text{eff}} \gets 0$ \Comment{turn off ascent near the end}
    \EndIf
    \State $x_{t-1} \gets x_{t-1} + \gamma_t^{\text{eff}} \nabla_{x_t} \big\|m\odot \big(y - \mathcal{A}(\hat{x}_0)\big)\big\|_2^2 + \sigma_t \varepsilon$
\EndFor
\State $\hat{x} \gets \hat{x}_0(x_0)$
\State Compute $d_{\text{meas}}$ using Eq.~\ref{eq:hvm_intrinsic}; if taxonomy constraints are not satisfied, resample $m$ and repeat
\State \Return intrinsic hallucinated image $\hat{x}$ and mask $m$
\end{algorithmic}
\end{algorithm}

\begin{algorithm}[t]
\caption{HalluGen -- Extrinsic Hallucination}
\label{alg:hallugen_extrinsic}
\begin{algorithmic}[1]
\Require Ground truth $x_{\text{gt}}$, measurement $y$, forward operator $\mathcal{A}(\cdot)$
\Require Diffusion model $\mu_{\theta}$, feature extractor $F(\cdot)$
\Require Step size $\{\lambda_t\}$, $\{\gamma_{1,t}\}$, $\{\gamma_{2,t}\}$
\Require Non-hallucinated baseline $x_{\text{base}}$
\Require Optional skip timestep $t_{\text{skip}}$ (or None), ascent cut-off timestep $t_{\text{stop}}$
\Require Interpolation flag $\text{interp} \in \{\text{true},\text{false}\}$, interpolation schedule $w_t$
\State Sample hallucination mask $m \in \{0,1\}^{H \times W}$ by entropy-based patch selection
\If{$t_{\text{skip}}$ is None}
    \State $t_{\text{start}} \gets T$
    \State Sample $x_T \sim \mathcal{N}(0, I)$
\Else
    \State $t_{\text{start}} \gets t_{\text{skip}}$
    \State Sample $x_{t_{\text{start}}} \sim q(x_{t_{\text{start}}} \mid x_{\text{base}})$ \Comment{forward process $x_{\text{base}}$}
\EndIf
\For{$t = t_{\text{start}}, t_{\text{start}}-1, \ldots, 1$}
    \State Compute Tweedie estimate $\hat{x}_0(x_t)$
    \If{$\text{interp} = \text{true}$}
        \State $\hat{x}_0(x_t) \gets (1 - m)\odot\big[w_t\,\hat{x}_0(x_t) + (1 - w_t)\,x_{\text{base}}\big] + m \odot \hat{x}_0(x_t)$
    \EndIf
    \State $x_{t-1} \gets \mu_{\theta}(x_t, t) - \lambda_t \nabla_{x_t} \big\|y - \mathcal{A}(\hat{x}_0)\big\|_2^2$
    \If{$t > t_{\text{stop}}$}
        \State $\gamma_{1,t}^{\text{eff}} \gets \gamma_{1,t}$,\quad $\gamma_{2,t}^{\text{eff}} \gets \gamma_{2,t}$
    \Else
        \State $\gamma_{1,t}^{\text{eff}} \gets 0$,\quad $\gamma_{2,t}^{\text{eff}} \gets 0$ \Comment{turn off ascent near the end}
    \EndIf
    \State $x_{t-1} \gets x_{t-1} + \gamma_{1,t}^{\text{eff}} \nabla_{x_t} \big\|m\odot (\hat{x}_0 - x_{\text{gt}})\big\|_2^2$
    \State $x_{t-1} \gets x_{t-1} + \gamma_{2,t}^{\text{eff}} \nabla_{x_t} \big\|m\odot (F(\hat{x}_0) - F(x_{\text{gt}}))\big\|_2^2 + \sigma_t \varepsilon$
\EndFor
\State $\hat{x} \gets \hat{x}_0(x_0)$
\State Compute $d_{\text{meas}}$ and $d_{\text{img}}$ using Eq.~\ref{eq:hvm_extrinsic}; if taxonomy constraints are not satisfied, resample $m$ and repeat
\State \Return extrinsic hallucinated image $\hat{x}$ and mask $m$
\end{algorithmic}
\end{algorithm}

\subsection{Ablations}
\label{subsec:hallugen_ablations}
In this section, we ablate each HalluGen component: HVM, entropy-based patch selection, and the feature-space loss.

\textbf{HVM maximizes hallucination while adhering to hallucination taxonomy.}
HVM enforces the intrinsic/extrinsic definitions in Equations~\ref{eq:intrinsic_definition} and \ref{eq:extrinsic_definition} by rejecting samples that do not exhibit sufficient deviation in masked regions. As shown in Tab.~\ref{tab:hvm}, enabling HVM increases measurement- and image-space losses for intrinsic cases by roughly 7\% and 20\%, indicating stronger violations of data fidelity as required. For extrinsic cases, HVM lowers measurement-space error while increasing image-space deviation by over 56\%, enforcing larger semantic changes while more strictly preserving measurement consistency. These results confirm that HVM improves taxonomy compliance by pushing intrinsic hallucinations toward greater measurement inconsistency and extrinsic hallucinations toward targeted semantic deviation under measurement constraints.

\textbf{Entropy-based patch selection and feature loss increase semantic deviations.}
The entropy-based patch selection module and the feature loss are designed to target semantically informative, hallucination-prone regions and to ensure that extrinsic hallucinations alter higher-level semantic content rather than only low-level pixel intensities. In Tab.\ref{tab:hvm}, removing entropy-based selection reduces both measurement- and image-space deviations for intrinsic cases, while enabling it increases these deviations by about 70\%, producing more localized and meaningful perturbations. For extrinsic cases, omitting the feature loss yields small deviations within the masked region. Including it more than doubles the image-space deviation while only slightly raising measurement-space error, which remains orders of magnitude lower than intrinsic cases. This shows that the feature loss strengthens higher-level semantic modification without compromising the strict measurement consistency required for extrinsic hallucinations.

\begin{table}[!hbpt]
\centering
\footnotesize
\caption{
\textbf{Effects of HVM, Entropy-based selection and Feature loss on hallucination taxonomy compliance using mean squared error within masked region (N=250).} 
}
\label{tab:hvm}
\begin{tabular}{lccccc}
\toprule
\textbf{Case} & \textbf{HVM} & \textbf{Ent.} &\textbf{Feat.} & \textbf{Meas.} & \textbf{Image} \\
\midrule
Intrinsic & x & x & x & $1.5 \times 10^{-3}$ & $3.8 \times 10^{-3}$ \\
Intrinsic & x & v & x & $2.5 \times 10^{-3}$ & $6.5 \times 10^{-3}$ \\
Intrinsic & v & v &x & $3.0 \times 10^{-3}$ & $7.0 \times 10^{-3}$\\
\midrule
Extrinsic & x & v & x & $1.6 \times 10^{-5}$ & $1.5 \times 10^{-3}$ \\
Extrinsic & x & x & v & $1.7 \times 10^{-5}$ & $2.7 \times 10^{-3}$ \\
Extrinsic & x & v & v & $2.6 \times 10^{-5}$ & $3.0 \times 10^{-3}$ \\
Extrinsic & v & v & v & $2.2 \times 10^{-5}$ & $4.7 \times 10^{-3}$ \\
\bottomrule
\end{tabular}
\vspace{-8pt} 
\end{table}

\subsection{Qualitative Results}
\label{subsec:hallugen_visual}
\begin{figure*}[t]
  \centering
  \includegraphics[width=0.98\linewidth]{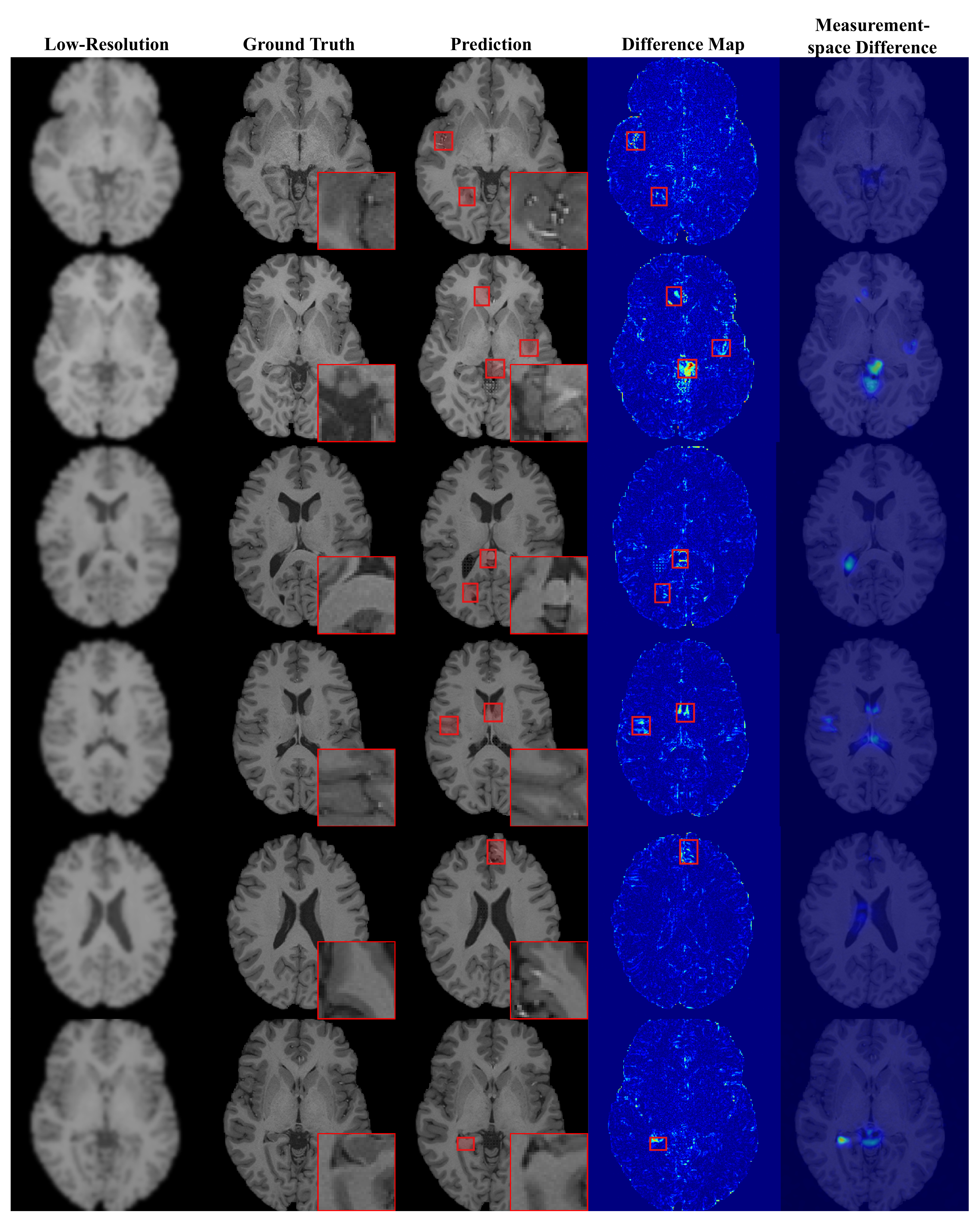}
  \caption{
    \textbf{Visual results of HalluGen examples on brain MRI.} Odd-numbered rows show intrinsic hallucinations, and even-numbered rows show extrinsic hallucinations.
  }
  \label{fig:hcp_total}
\end{figure*}

\begin{figure*}[t]
  \centering
  \includegraphics[width=0.98\linewidth]{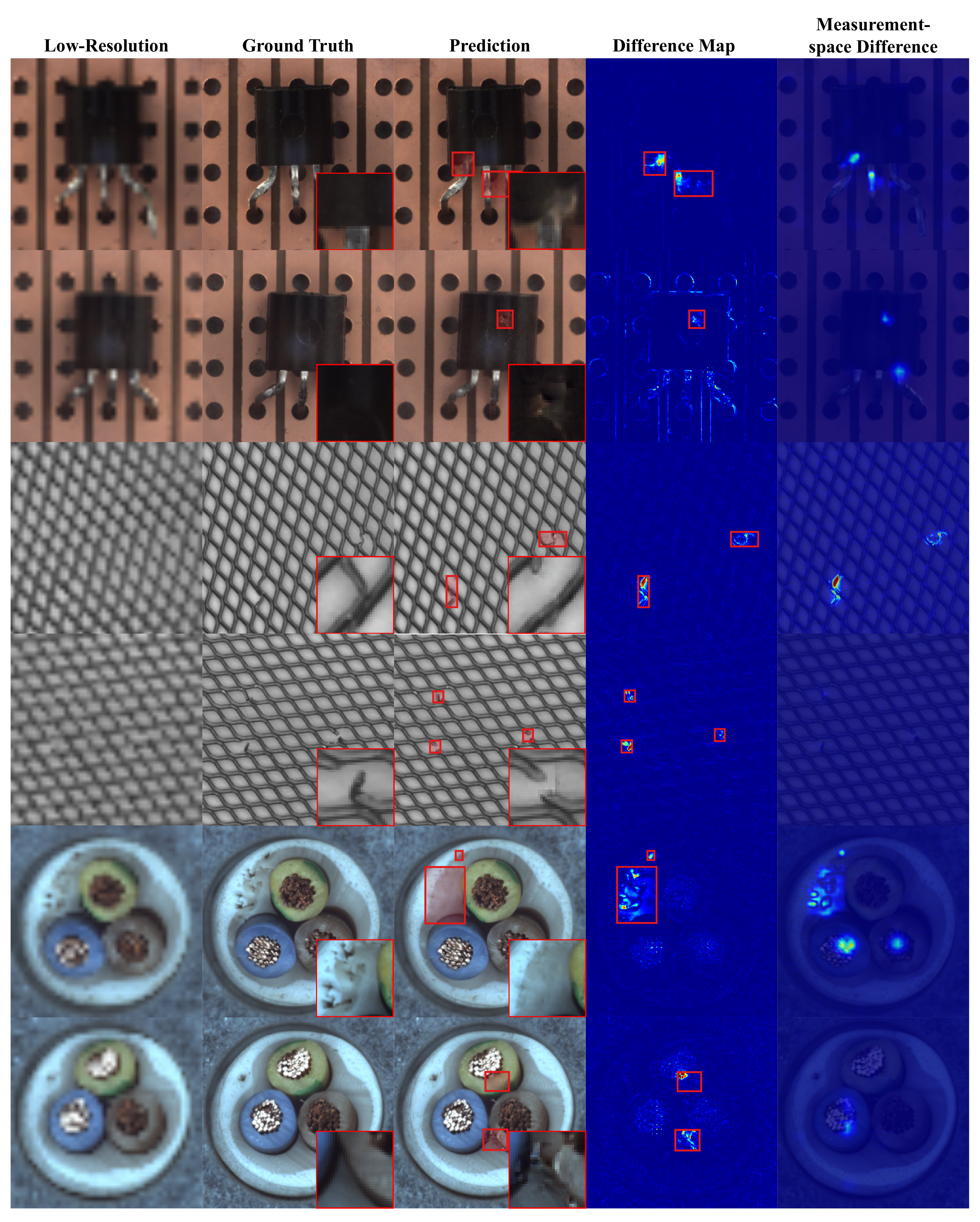}
  \caption{
    \textbf{Visual results of HalluGen examples on MVTec AD.} Odd-numbered rows show extrinsic hallucinations, and even-numbered rows show intrinsic hallucinations. Row 4 and 5 use ground truth mask from MVTec AD to show the flexibility of HalluGen with different size and shape of masks.
  }
  \label{fig:mvtec_total}
\end{figure*}

\begin{figure*}[t]
  \centering
  \includegraphics[width=0.98\linewidth]{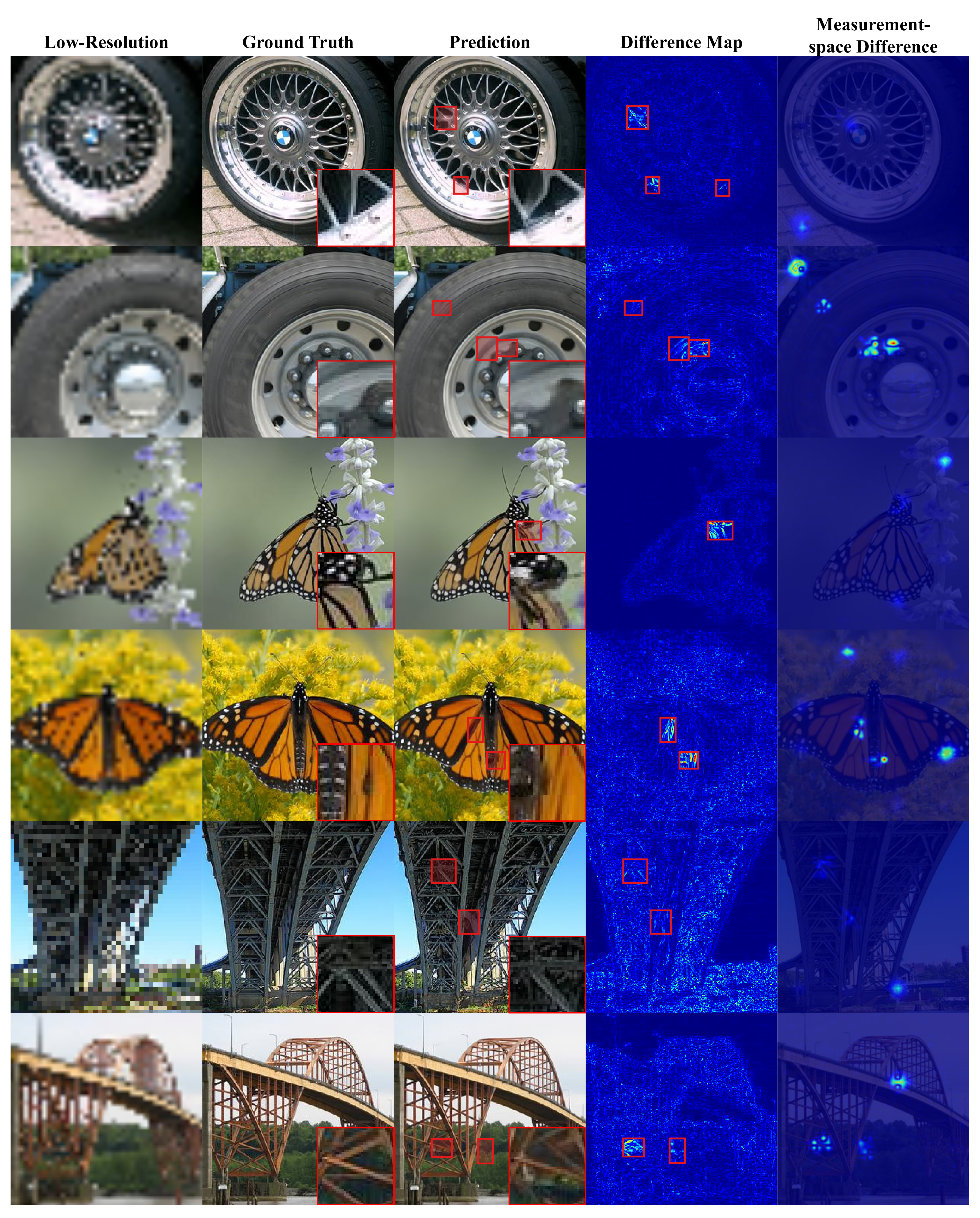}
  \caption{
    \textbf{Visual results of HalluGen examples on ImageNet.} Odd-numbered rows show extrinsic hallucinations, and even-numbered rows show intrinsic hallucinations.
  }
  \label{fig:imagenet_total}
\end{figure*}

We provide additional visual examples in Fig.~\ref{fig:hcp_total}, Fig.~\ref{fig:mvtec_total}, and Fig.~\ref{fig:imagenet_total}, corresponding to MRI, industrial, and natural images, respectively. For natural images, we initialize sampling from the ground-truth image rather than $x_{\text{base}}$ to preserve fidelity outside the hallucination mask and clearly isolate the injected content. These results demonstrate that HalluGen generalizes across diverse imaging domains without modification.
 
\subsection{Failure Case Analysis}
\label{subsec:hallugen_failure}
We observed three recurring challenges during dataset construction with HalluGen.

\textbf{(1) Sensitivity to regional texture and homogeneity.}
Hallucination severity depends on the local structure of the selected region. When masks cover smooth or homogeneous areas, the diffusion prior can dominate and suppress the gradient-ascent signal needed to induce hallucination. For extrinsic cases, such regions also shrink the effective null space of the forward operator $\mathcal{A}(\cdot)$, making semantic deviations harder to produce. These issues are largely mitigated by HVM and entropy-based patch selection, which favor more informative regions.

\textbf{(2) Limited control over precise morphological changes.}
Because HalluGen relies on stochastic sampling, it cannot deterministically specify the exact morphology or topology of hallucinated structures. While HalluGen provides spatial control and severity modulation, it does not guarantee identical hallucination patterns across runs, suggesting future work on structure-aware hallucination priors.

\textbf{(3) Diffusion prior strongly influences hallucination realism.}
Hallucination realism depends on the data distribution of the diffusion prior. To study this, we trained two MVTec AD diffusion models: (i) class-specific and (ii) multi-class. With identical hyperparameters, the class-specific prior produced more coherent hallucinations, whereas the multi-class prior often generated implausible artifacts. Strong gradients can push samples across object manifolds when the prior spans diverse categories. Lower gradient strengths help stabilize results, but adaptive or class-aware gradient schedules remain an open direction.

\section{Dataset}
\subsection{Instructions for Dataset}
\label{subsec:dataset_instruction}
The dataset is constructed for low-field MRI enhancement. We first pre-train a diffusion model using axial slices from 1{,}000 HCP scans and simulate low-field MRI using the forward model from DynamicDPS~\cite{dynamicdps}:

\begin{equation}
\mathcal{A}(x) = \text{Blur}\left(\text{DS}_{k}\left(\Gamma{\gamma}(x)\right)\right),
\qquad k = 4,\ \gamma = 0.7,
\label{eq:lf_forward}
\end{equation}

where $\Gamma_{\gamma}$ is a gamma transform, $\text{DS}_{k}$ is $k$-fold downsampling, and \text{Blur} applies spatial smoothing. These parameters create a moderately ill-posed setting—severe degradation leads to uncontrolled global hallucinations, while mild degradation limits the null space needed for extrinsic cases.

We then generate 1{,}450 non-hallucinated predictions using DPS~\cite{chung2023diffusion}. These serve as (i) the baseline for timestep skipping and (ii) the reference for interpolation outside the hallucination mask, ensuring hallucinations remain localized. Applying HalluGen with intrinsic and extrinsic settings produces 1{,}450 samples for each category, resulting in a dataset of 4{,}350 images in total.

The dataset consists of following:
For each ground truth image $x_{gt}$:
\begin{itemize}[leftmargin=1.5em]
    \item file ID and slice index for ground truth
    \item measurement image 
    \item non-hallucinated baseline from DPS~\cite{chung2023diffusion}
    \item intrinsic and extrinsic hallucinations from HalluGen
    \item binary hallucination mask for each type
\end{itemize}

All data are stored as NumPy arrays (.npy).

\subsection{Computational Cost} 
\label{subsec:dataset_computational}
We used $1 \times$ A6000 (48GB VRAM) GPU to generate the dataset.
It takes approximately $60s$ to generate intrinsic and $90s$ to generate extrinsic sample with batch size 1.
However, since HVM may reject the sample with insufficient hallucination, it can take slightly longer.

\subsection{Limitations}
\label{subsec:dataset_limitaiton}
The current dataset includes only healthy adult brain MRI scans, as the diffusion prior was trained on the HCP~\cite{hcp}. As a result, the hallucination characteristics may not fully reflect those that occur in pathological or clinically diverse populations. Nonetheless, this release serves as an initial foundation for hallucination analysis in image restoration, providing the first controlled dataset of intrinsic and extrinsic hallucinations.

\subsection{License and Ethics Statement}
\label{subsec:dataset_license}
This work makes use of human brain MRI data provided by the Human Connectome Project (HCP)~\cite{hcp}. Our dataset includes ground-truth MR images obtained from HCP and synthetic hallucination outputs generated using HalluGen. All HCP images are distributed only to users who have agreed to the HCP Open Access Data Use Terms\footnote{\url{https://www.humanconnectome.org/study/hcp-young-adult/document/wu-minn-hcp-consortium-open-access-data-use-terms}}. Users must obtain access to the original HCP data directly from the HCP platform; we do not redistribute any HCP imaging data and do not grant additional usage rights beyond those provided by HCP.

All hallucination masks and synthetic hallucination images produced by HalluGen are released under a CC BY-NC 4.0 International License\footnote{\url{https://creativecommons.org/licenses/by-nc/4.0/}}, permitting non-commercial research use with attribution. Our source code is released under the MIT License\footnote{\url{https://opensource.org/licenses/MIT}}. The dataset is intended solely for academic research and must not be used for clinical decision-making or deployed in medical settings.

No additional human subjects were recruited or scanned for this work. Our use complies with the HCP ethics and data protection policies.

\section{SHAFE}
\subsection{Implementation Details}
\label{subsec:shafe_details}
We provide additional implementation details for SHAFE, illustrated in Algo.~\ref{alg:shafe}. SHAFE is a full-reference metric that takes as input the predicted reconstruction $\hat{x}$ and the ground-truth image $x_{\text{gt}}$. Both images are first passed through a low-pass filter to suppress high-frequency artifacts that do not correspond to hallucinations (e.g., checkerboard patterns), thereby reducing false positives.

Feature representations are extracted using a pre-trained encoder (e.g., DINOv3). We use multi shallow-layer features to avoid semantic bias from deeper layers and to emphasize structural and textural cues that better reflect hallucination-related deviations. Cosine distances are computed patch-wise between the feature maps of $\hat{x}$ and $x_{\text{gt}}$.

Traditional image-quality metrics that aggregate patch scores uniformly (e.g., SSIM, LPIPS), but this is suboptimal for hallucination assessment, as hallucinations tend to be sparse and highly localized (see Fig.~\ref{fig:localization}). To address this, inspired by soft-attention pooling~\cite{attentionpooling}, we apply a weighted softmax aggregation over patch-level distances, enabling SHAFE to upweight semantically abnormal regions while downweighting benign variations.

\textbf{Hyperparamters.} We use the following settings:
\begin{itemize}[leftmargin=1.5em]
    \item Low-pass filter radius: 50
    \item Feature backbone: DINOv3~\cite{dinov3}, ResNet50~\cite{resnet50aa}, MedSAM~\cite{medsam}
    \item Feature layer indices: [1, 2, 3], [1, 2], [-1]
    \item $tau:$ 0.01, 0.01, 0.005
\end{itemize}

\begin{algorithm}[!htbp]
\caption{SHAFE: Semantic Hallucination Assessment via Feature Evaluation}
\label{alg:shafe}
\begin{algorithmic}[1]
\Require Prediction $\hat{x}$, ground truth $x_{\text{gt}}$
\Require Low-pass filter $\mathrm{LP}(\cdot)$, pretrained encoder $\{f_\ell\}_{\ell \in \mathcal{L}}$ (e.g., DINOv3)
\Require Temperature $\tau > 0$
\Statex
\State $\tilde{x} \gets \mathrm{LP}(\hat{x}), \quad \tilde{x}_{\text{gt}} \gets \mathrm{LP}(x_{\text{gt}})$
\For{each layer $\ell \in \mathcal{L}$}
    \State $F_\ell \gets f_\ell(\tilde{x}), \quad G_\ell \gets f_\ell(\tilde{x}_{\text{gt}})$
\EndFor
\State $F \gets \mathrm{Concat}\big(\{F_\ell\}_{\ell \in \mathcal{L}}\big), \quad G \gets \mathrm{Concat}\big(\{G_\ell\}_{\ell \in \mathcal{L}}\big)$
\State Reshape $F, G$ into patch features $\{F_i\}_{i=1}^N$, $\{G_i\}_{i=1}^N$
\For{$i = 1$ to $N$}
    \State $\delta_{\mathrm{cos}, i} \gets 1 - \dfrac{\langle F_i, G_i \rangle}{\|F_i\|_2 \, \|G_i\|_2}$ \Comment{cosine distance per patch}
\EndFor
\State $w_i \gets \dfrac{\exp(\delta_{\mathrm{cos}, i}/\tau)}{\sum_{j=1}^{N} \exp(\delta_{\mathrm{cos}, j}/\tau)} \quad \text{for } i = 1,\dots,N$
\State $\mathrm{SHAFE} \gets \sum_{i=1}^{N} w_i \, \delta_{\mathrm{cos}, i}$
\State \Return $\mathrm{SHAFE}$ \Comment{optionally also return patch map $\{w_i \delta_{\mathrm{cos}, i}\}$}
\end{algorithmic}
\end{algorithm}

\subsection{Ablations}
\label{subsec:shafe_ablation}
We ablate the components of SHAFE, including the low-pass (LP) filter, weighted softmax aggregation, temperature~$\tau$, and feature-layer selection. All experiments use a subset of our HalluGen dataset (N=300) for both intrinsic and extrinsic hallucinations.

\paragraph{Effect of Low-Pass Filtering and Weighted-softmax Aggregation.}
Table~\ref{tab:shafe} shows that both LP filtering and weighted softmax aggregation improve hallucination detection (AUC). LP filtering suppresses high-frequency noise, helping SHAFE emphasize coherent structural differences. Weighted softmax aggregation further highlights patches with stronger feature deviations, and already outperforms uniform pooling, reflecting the spatially localized nature of hallucinations. Combining both modules increases AUC from 0.52 to 0.78, demonstrating the complementary benefits of frequency smoothing and adaptive spatial weighting.

\begin{table}[!hbpt]
\centering
\footnotesize
\caption{
\textbf{Ablation of low-pass filtering (LP) and weighted-softmax aggregation in SHAFE, evaluated using AUC on hallucination detection (N=300).}
Both components improve detection sensitivity, and their combination yields the highest performance, indicating that suppressing high-frequency noise while adaptively weighting salient patches is crucial for detecting hallucinations.
}

\label{tab:shafe}
\begin{tabular}{cccc}
\toprule
\textbf{LP} &\textbf{Weighted softmax} & \textbf{AUC} \\
\midrule
 x & x & 0.52 \\
 v & x &  0.55 \\
 x & v &  0.64 \\
 v & v &  \textbf{0.78}\\
\bottomrule
\end{tabular}
\end{table}

\begin{figure}[t]
  \centering
  \includegraphics[width=\columnwidth]{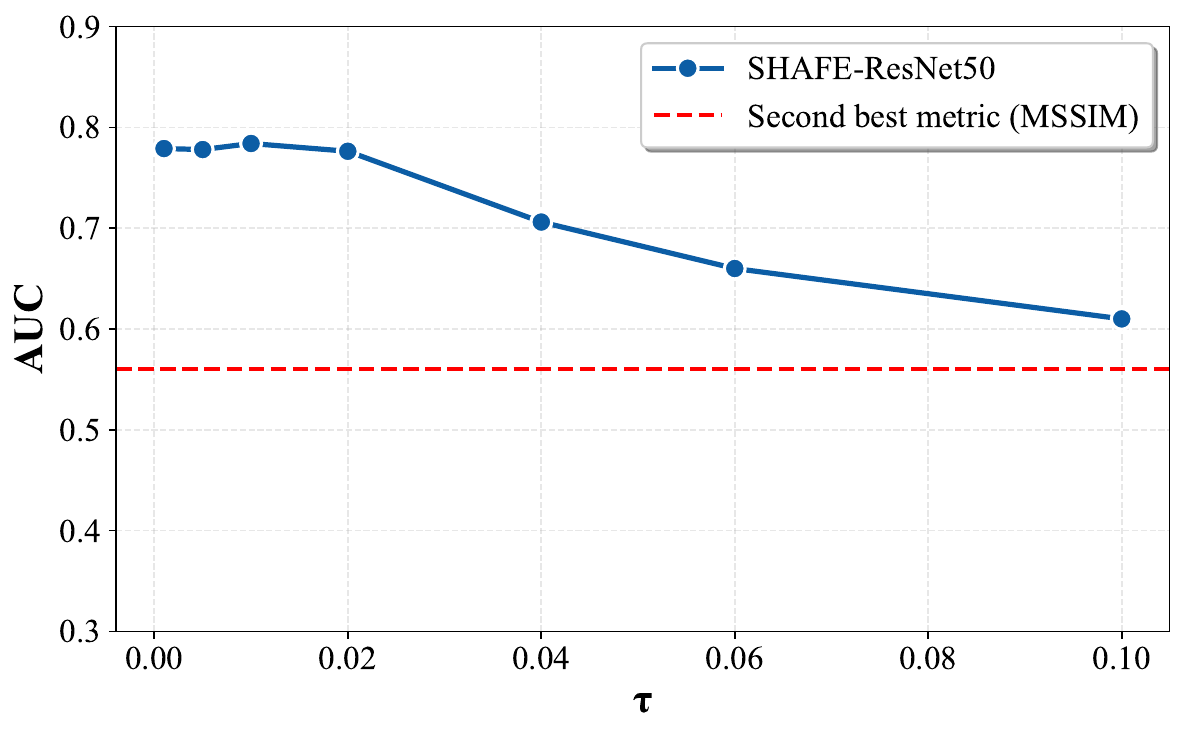}
  \caption{
\textbf{Effect of temperature $\tau$ in softmax aggregation for SHAFE (AUC vs.\ $\tau$).}
Small temperatures approximate max-pooling and yield strong performance, stable until $\tau \approx 0.02$. Larger values smooth patch weighting and gradually reduce discriminability, yet SHAFE remains superior to the second-best metric across all tested values.
}
\label{fig:tau_effect}
\end{figure}

\paragraph{Effect of Temperature $\tau$.}
Figure~\ref{fig:tau_effect} illustrates the impact of varying the temperature $\tau$ in the softmax aggregation. Very small values (e.g., $\tau \leq 0.001$) approximate max-pooling and already produce strong AUC, indicating that hallucinations often manifest as localized feature deviations. Performance remains stable until $\tau \approx 0.02$, after which AUC decreases gradually as the softmax distribution becomes more uniform and less discriminative. Nonetheless, SHAFE continues to outperform the second-best baseline across all tested values, highlighting robustness to $\tau$.

\paragraph{Impact of Feature-Layer Selection.}
Table~\ref{tab:feature} reports the effect of feature-layer selection when using SHAFE-ResNet50. Using only deeper layers degrades performance, likely due to dataset-specific and task-specific bias that do not fully align with medical and industrial domains. Shallow features capture texture and low-level structure, which are more sensitive to fine-grained hallucinations. The best performance is achieved when combining multiple shallow layers (AUC=0.78), supporting our design choice that multi-layer early features provide richer representations for hallucination scoring.

\begin{table}[!hbpt]
\centering
\footnotesize
\caption{
\textbf{Impact of feature-layer selection on SHAFE hallucination detection (AUC, N=300) using SHAFE-ResNet50.}
Deep layers alone degrade performance due to semantic bias from ImageNet pretraining, whereas combining shallow layers provides the highest AUC, supporting the use of multi-layer early features for hallucination sensitivity.
}

\label{tab:feature}
\begin{tabular}{cccc}
\toprule
\textbf{Layer 1} &\textbf{Layer 2} & \textbf{Layer 3} & \textbf{AUC}\\
\midrule
 v & x & x & 0.75 \\
 x & v & x & 0.76 \\
 x & x & v & 0.63 \\
 v & v & x & \textbf{0.78}\\
 v & v &  v & 0.72\\
\bottomrule
\end{tabular}
\vspace{-8pt} 
\end{table}

\subsection{Localization}
\label{subsec:shafe_localization}
\begin{figure}[t]
  \centering
  \includegraphics[width=\columnwidth]{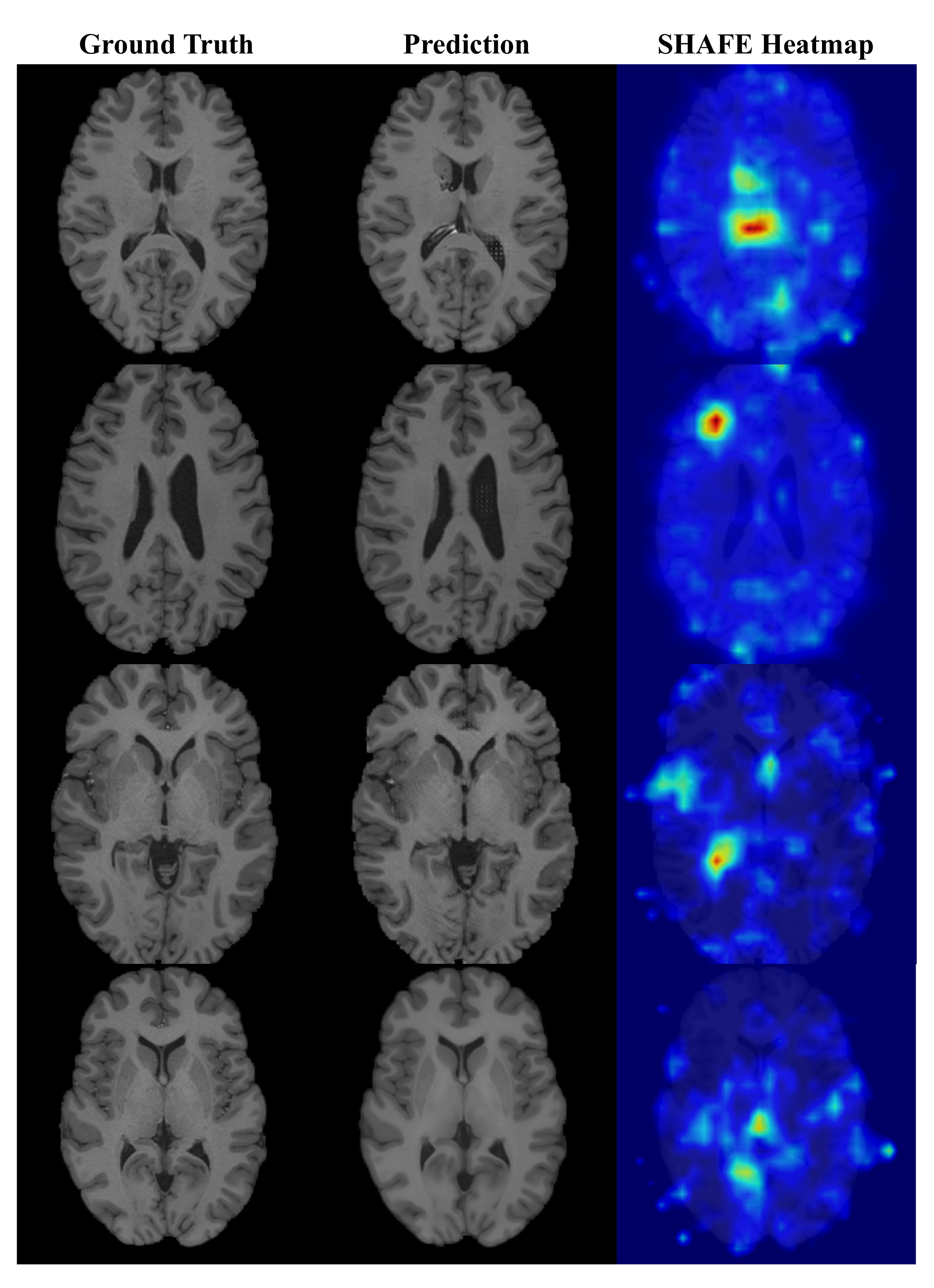}
  \caption{
    \textbf{Visual results of SHAFE heatmap on both synthetic (HalluGen) and real restoration outputs.} First and second rows are extrinsic and intrinsic examples from HalluGen and third and fourth are real restoration outputs from ESRGAN and SwinIR.
  }
  \label{fig:localization}
\vspace{-8pt} 
\end{figure}

Fig.~\ref{fig:localization} presents the localization performance of SHAFE on both HalluGen samples and real restoration outputs generated by ESRGAN~\cite{ESRGAN} and SwinIR~\cite{swinir}. SHAFE consistently highlights hallucinated regions such as distorted lateral ventricles and disrupted sulcal boundaries, demonstrating its ability to detect both synthetic and naturally occurring failures. The resulting heatmaps are characteristically sparse, emphasizing localized deviations rather than diffuse artifacts. This sparsity further justifies the use of weighted softmax aggregation over uniform pooling, as it enables SHAFE to selectively amplify subtle but semantically meaningful hallucinated features.

\subsection{Limitations of SHAFE}
\label{subsec:shafe_failure}
\begin{figure}[t]
  \centering
  \includegraphics[width=\columnwidth]{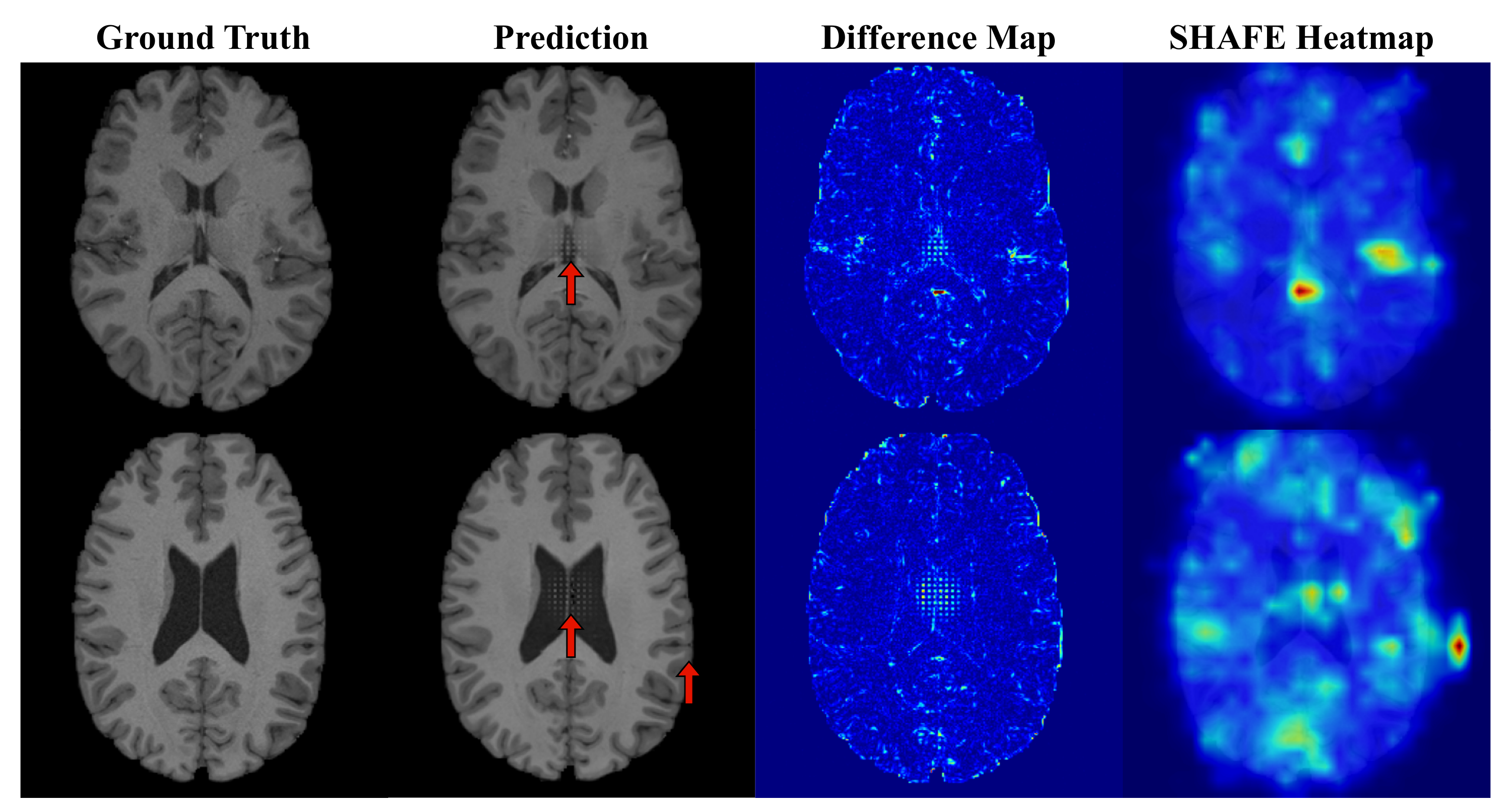}
\caption{
\textbf{Visual examples of SHAFE limitations (red arrows).} SHAFE is less sensitive to high-frequency reconstruction artifacts (e.g., periodic grid patterns) due to the use of low-pass filtering, which suppresses non-semantic noise. In addition, SHAFE may highlight boundary regions where intensity differences are large but carry limited semantic relevance.
}
  \label{fig:failure}
\vspace{-8pt} 
\end{figure}

Fig.~\ref{fig:failure} illustrates representative limitations of SHAFE. First, SHAFE is less sensitive to high-frequency artifacts (e.g., periodic grid patterns) due to the low-pass filtering step, which suppresses high-frequency content because hallucinations in our taxonomy mainly occur in the low- to mid-frequency range. Consequently, images with such artifacts may obtain SHAFE scores similar to clean ones. Second, SHAFE can assign high responses to large intensity differences in semantically uninformative regions, such as brain boundaries, where intensity mismatches are high but structural deviation is minimal.

\subsection{Inference Speed}
\label{subsec:shafe_computational}
\begin{figure}[!hbpt]
  \centering
  \includegraphics[width=\columnwidth]{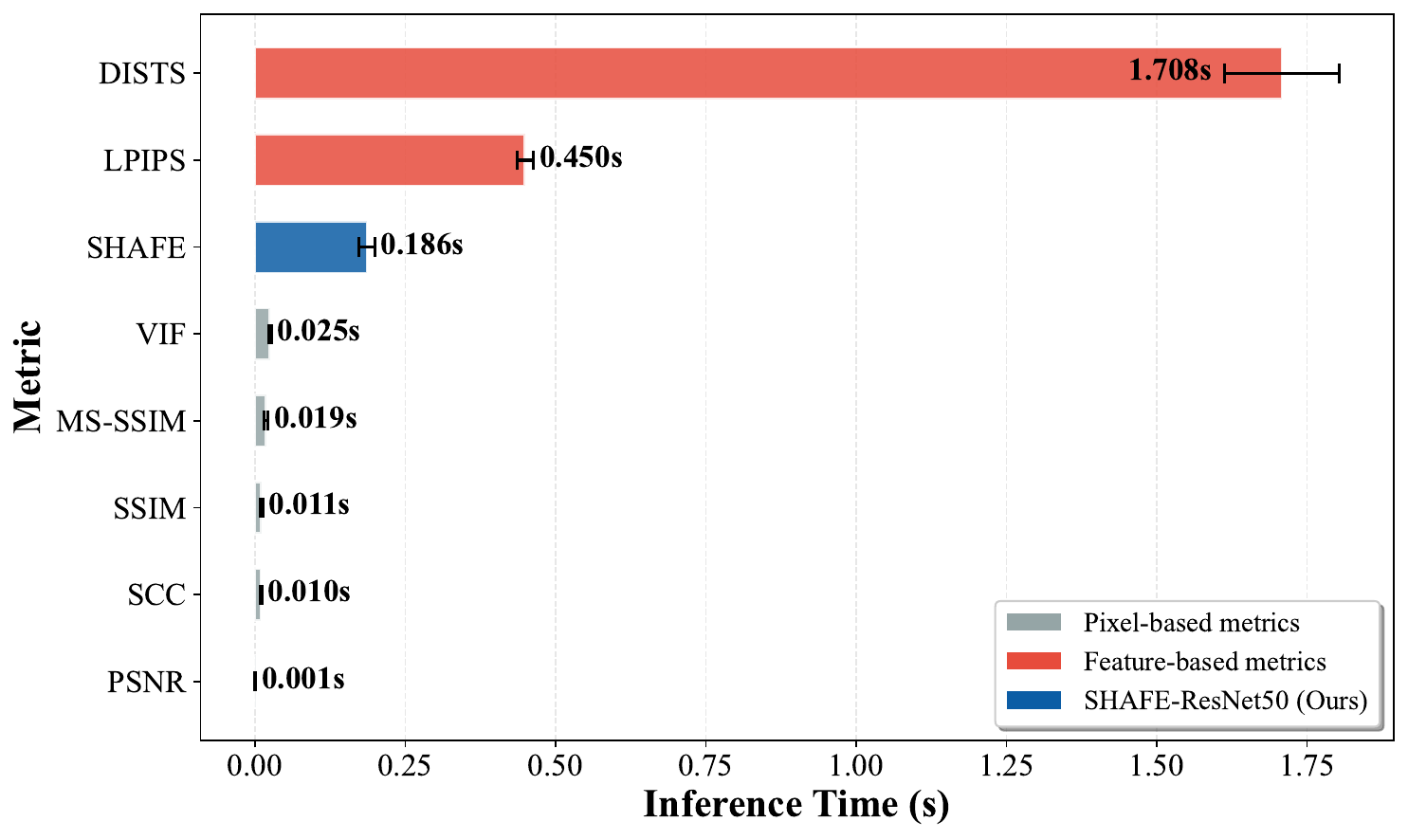}
\caption{
\textbf{Inference time comparison for baseline metrics vs. SHAFE-ResNet50.}
Despite using feature extraction and weighted softmax, SHAFE remains computationally efficient, running significantly faster than other feature-based metrics such as LPIPS and DISTS. This demonstrates that SHAFE achieves superior detection performance without incurring substantial computational overhead. AMD Ryzen 7 CPU was used for testing.
}
\label{fig:inf_time}
\vspace{-8pt} 
\end{figure}

Fig.~\ref{fig:inf_time} compares the inference time of SHAFE-ResNet50 with baseline metrics in our hallucination benchmark. While SHAFE is naturally slower than pixel-based metrics such as PSNR and SSIM, it remains significantly more efficient than feature-based alternatives: SHAFE is about 2.4$\times$ faster than LPIPS and 9$\times$ faster than DISTS. This shows that the combination of low-pass filtering, shallow feature extraction, and weighted softmax aggregation provides strong detection performance without the heavy computational cost of deep feature metrics. All timings were measured on a local machine with an AMD Ryzen 7 CPU.

\section{Reference-free Hallucination Detector}
\label{sec:hallu_detector}
\begin{figure}[!htbp]
  \centering
  \includegraphics[width=\columnwidth]{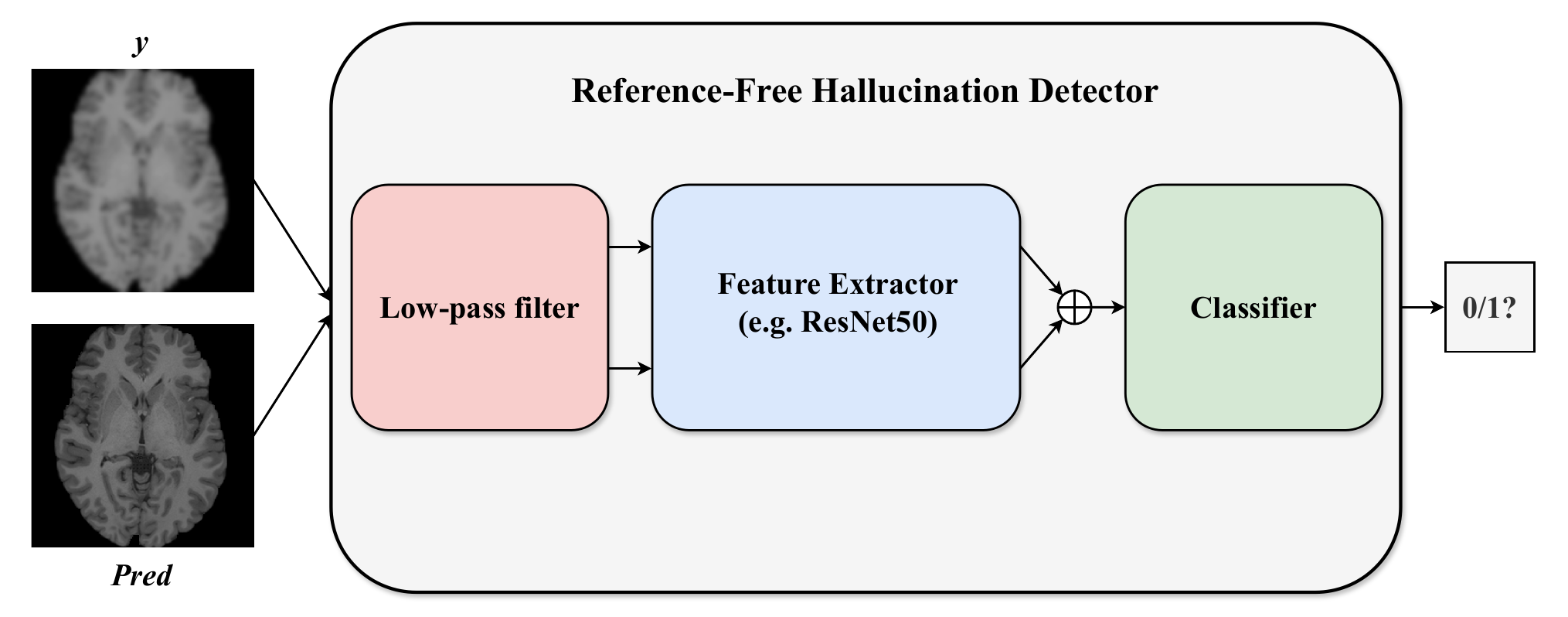}
  \caption{
    \textbf{Network architecture of reference-free hallucination detector.} Prediction and measurement images are fed into the model, applied low-pass filter to remove high-frequency noise and then classified using a simple CNN architecture such as ResNet50.
  }
  \label{fig:detector}
\end{figure}

We describe the architecture of our reference-free hallucination detector. As shown in Fig.~\ref{fig:detector}, the detector takes the measurement $y$ and predicted reconstruction $\hat{x}$ as inputs, both of which are first low-pass filtered to suppress high-frequency noise. The filtered images are passed through a shallow ResNet-50 feature extractor, and their features are concatenated before a binary classification head. We also experimented with concatenating the raw inputs prior to feature extraction, but this caused the model to focus on low-level differences (e.g., small shifts or texture changes) rather than true semantic discrepancies. The detector is trained with cross-entropy loss for 100 epochs using a learning rate of 0.001 on HalluGen-generated samples. Because it relies only on the measurement–prediction pair, the detector supports fully reference-free hallucination detection.

%% file: main.bib
@String(CVPR= {IEEE Conf. Comput. Vis. Pattern Recog.})

@String(ICCV= {Int. Conf. Comput. Vis.})

@String(ECCV= {Eur. Conf. Comput. Vis.})

@String(CVPR  = {CVPR})

@String(ICCV  = {ICCV})

@String(ECCV  = {ECCV})

@inproceedings{ddpm,
 author = {Ho, Jonathan and Jain, Ajay and Abbeel, Pieter},
 booktitle = {Advances in Neural Information Processing Systems},
 title = {Denoising Diffusion Probabilistic Models},
 year = {2020}
}

@inproceedings{ddrm,
    title={Denoising Diffusion Restoration Models},
    author={Bahjat Kawar and Michael Elad and Stefano Ermon and Jiaming Song},
    booktitle={Advances in Neural Information Processing Systems},
    year={2022}
}

@article{diffusioniqt,
  title={A 3D Conditional Diffusion Model for Image Quality Transfer--An Application to Low-Field MRI},
  author={Kim, Seunghoi and Tregidgo, Henry FJ and Eldaly, Ahmed K and Figini, Matteo and Alexander, Daniel C},
  journal={arXiv preprint arXiv:2311.06631},
  year={2023}
}

@article{zeroddrm,
title={Zero-Shot Image Restoration Using Denoising Diffusion Null-Space Model},
author={Wang, Yinhuai and Yu, Jiwen and Zhang, Jian},
journal={The Eleventh International Conference on Learning Representations},
year={2023}
}

@InProceedings{chung2023solving,
  title={Solving 3D Inverse Problems using Pre-trained 2D Diffusion Models},
  author={Chung, Hyungjin and Ryu, Dohoon and McCann, Michael T and Klasky, Marc L and Ye, Jong Chul},
  booktitle={IEEE/CVF Conference on Computer Vision and Pattern Recognition},
  year={2023}
}

@InProceedings{localdiff,
author="Kim, Seunghoi
and Jin, Chen
and Diethe, Tom
and Figini, Matteo
and Tregidgo, Henry F. J.
and others",
title="Tackling Structural Hallucination in Image Translation with Local Diffusion",
booktitle="European Conference on Computer Vision (ECCV)",
year="2024",
}

@inproceedings{
chung2023diffusion,
title={Diffusion Posterior Sampling for General Noisy Inverse Problems},
author={Hyungjin Chung and Jeongsol Kim and Michael Thompson Mccann and Marc Louis Klasky and Jong Chul Ye},
booktitle={The Eleventh International Conference on Learning Representations },
year={2023},
}

@inproceedings{diffpir, 
      title={Denoising Diffusion Models for Plug-and-Play Image Restoration},
      author={Yuanzhi Zhu and Kai Zhang and Jingyun Liang and Jiezhang Cao and Bihan Wen and others},
      booktitle={IEEE Conference on Computer Vision and Pattern Recognition Workshops (NTIRE)},
      year={2023},
}

@misc{stsl,
      title={Beyond First-Order Tweedie: Solving Inverse Problems using Latent Diffusion}, 
      author={Rout, L and Chen, Y and Kumar, A and Caramanis, C and Shakkottai, S and Chu, W},      
      journal={IEEE/CVF Conference on Computer Vision and Pattern Recognition (CVPR)},
      year={2024}
}

@misc{daps,
      title={Improving Diffusion Inverse Problem Solving with Decoupled Noise Annealing}, 
      author={Bingliang Zhang and Wenda Chu and Julius Berner and Chenlin Meng and Anima Anandkumar and Yang Song},
      year={2024},
      eprint={2407.01521}, 
}

@Article{brats,
  author = {Menze, B. H. and Jakab, A. and Bauer, S. and others},
  title = {The Multimodal Brain Tumor Image Segmentation Benchmark (BRATS)},
  journal = {IEEE Transactions on Medical Imaging},
  year = {2015},
}

@article{hcp,
  title={Advances in diffusion MRI acquisition and processing in the Human Connectome Project},
  author={Stamatios N. Sotiropoulos and Sa{\^a}d Jbabdi and Junqian Xu and Jesper L. R. Andersson and Steen Moeller and others},
  journal={NeuroImage},
  year={2013},
  volume={80},
  pages={125-143}
}

@InProceedings{ESRGAN,
    author = {Wang, Xintao and Yu, Ke and Wu, Shixiang and Gu, Jinjin and Liu, Yihao and others},
    title = {ESRGAN: Enhanced super-resolution generative adversarial networks},
    booktitle = {The European Conference on Computer Vision Workshops (ECCVW)},
    year = {2018}
}

@article{iqt_pio,
    author = {Alexander, Daniel C. and Darko, Zikic and Ghosh,  Aurobrata and Tanno, Ryutaro and Wottschel, Viktor and others},
    title = {Image quality transfer and applications in diffusion MRI},
    journal = {NeuroImage},
    year = {2017},
    volume = {152},
    pages = {283--298},
}

@article{iqt_stochastic,
  title = {Low-field magnetic resonance image enhancement via stochastic image quality transfer}, 
  journal = {Medical Image Analysis},
  volume = {87},
  pages = {102807},
  year = {2023},
  issn = {1361-8415},
  author = {Hongxiang Lin and Matteo Figini and Felice D'Arco and Godwin Ogbole and Ryutaro Tanno and others},
}

@article{synthsr,
  author = {Juan E. Iglesias  and Benjamin Billot  and Yaël Balbastre  and Colin Magdamo  and Steven E. Arnold and others},
  title = {SynthSR: A public AI tool to turn heterogeneous clinical brain scans into high-resolution T1-weighted images for 3D morphometry},
  journal = {Science Advances},
  volume = {9},
  number = {5},
  year = {2023},
}

@article{uliqt,
author = {Lau, Vick and Xiao, Linfang and Zhao, Yujiao and Su, Shi and Ding, Ye and others}, 
title = {Pushing the limits of low-cost ultra-low-field MRI by dual-acquisition deep learning 3D superresolution},
journal = {Magnetic Resonance in Medicine},
volume = {90},
number = {2},
pages = {400-416},
year = {2023}
}

@article{reconstructanything,
  title={Reconstruct Anything Model: a lightweight foundational model for computational imaging},
  author={Terris, Matthieu and Hurault, Samuel and Song, Maxime and Tachella, Julián},
  journal={arXiv preprint arXiv:2503.08915},
  year={2025}
}

@InProceedings{dynamicdps,
        author = { Kim, Seunghoi AND Tregidgo, Henry F. J. AND Figini, Matteo AND Jin, Chen AND Joshi, Sarang AND Alexander, Daniel C.},
        title = { { Tackling Hallucination from Conditional Models for Medical Image Reconstruction with DynamicDPS } },
        booktitle = {proceedings of Medical Image Computing and Computer Assisted Intervention -- MICCAI},
        year = {2025},
}

@article{hallucination_tomo,
  author    = {Bhadra, S. and Kelkar, V. A. and Brooks, F. J. and Anastasio, M. A.},
  title     = {On Hallucinations in Tomographic Image Reconstruction},
  journal   = {IEEE Transactions on Medical Imaging},
  year      = {2021},
  volume    = {40},
  number    = {11},
  pages     = {3249--3260},
}

@InProceedings{hallucination_metric,
        author = { Tivnan, Matthew and Yoon, Siyeop and Chen, Zhennong and Li, Xiang and Wu, Dufan and others},
        title = { { Hallucination Index: An Image Quality Metric for Generative Reconstruction Models } },
        booktitle = {proceedings of Medical Image Computing and Computer Assisted Intervention -- MICCAI},
        year = {2024},
}

@inproceedings{lpips,
  title={The Unreasonable Effectiveness of Deep Features as a Perceptual Metric},
  author={Zhang, Richard and Isola, Phillip and Efros, Alexei A and Shechtman, Eli and Wang, Oliver},
  booktitle={Proceedings of the IEEE/CVF Conference on Computer Vision and Pattern Recognition (CVPR)},
  year={2018}
}

@article{synth_survey,
    author = {Gopinath, Karthik and Hoopes, Andrew and Alexander, Daniel C. and Arnold, Steven E. and Balbastre, Yael and others},
    title = {Synthetic data in generalizable, learning-based neuroimaging},
    journal = {Imaging Neuroscience},
    volume = {2},
    pages = {1-22},
    year = {2024},
}

@inproceedings{
modeinterpolation,
title={Understanding Hallucinations in Diffusion Models through Mode Interpolation},
author={Sumukh K Aithal and Pratyush Maini and Zachary Chase Lipton and J Zico Kolter},
booktitle={The Thirty-eighth Annual Conference on Neural Information Processing Systems},
year={2024}
}

@article{videohallucer,
    title={VideoHallucer: Evaluating Intrinsic and Extrinsic Hallucinations in Large Video-Language Models},
    author={Wang, Yuxuan and Wang, Yueqian and Zhao, Dongyan and Xie, Cihang and Zheng, Zilong},
    journal={arxiv},
    year={2024}
}

@inproceedings{SRCNN,
  title={Learning a Deep Convolutional Network for Image Super-Resolution},
  author={Chao Dong and Chen Change Loy and Kaiming He and Xiaoou Tang},
  booktitle={European Conference on Computer Vision (ECCV)},
  year={2014},
}

@InProceedings{sr_transformer,
    author    = {Chen, Xiangyu and Wang, Xintao and Zhou, Jiantao and Qiao, Yu and Dong, Chao},
    title     = {Activating More Pixels in Image Super-Resolution Transformer},
    booktitle = {Proceedings of the IEEE/CVF Conference on Computer Vision and Pattern Recognition (CVPR)},
    year      = {2023},
}

@InProceedings{edsr,
  author = {Lim, Bee and Son, Sanghyun and Kim, Heewon and Nah, Seungjun and Lee, Kyoung Mu},
  title = {Enhanced Deep Residual Networks for Single Image Super-Resolution},
  booktitle = {The IEEE Conference on Computer Vision and Pattern Recognition (CVPR) Workshops},
  year = {2017}
}

@article{swinir,
  title={SwinIR: Image Restoration Using Swin Transformer},
  author={Liang, Jingyun and Cao, Jiezhang and Sun, Guolei and Zhang, Kai and Van Gool, Luc and Timofte, Radu},
  journal={arXiv preprint arXiv:2108.10257},
  year={2021}
}

@inproceedings{perceptiontradeoff,
  author       = {Yochai Blau and
                  Tomer Michaeli},
  title        = {The Perception-Distortion Tradeoff},
  booktitle    = {The IEEE / CVF Computer Vision and Pattern Recognition Conference (CVPR)},
  year         = {2018},
}

@article{dists,
  title={Image Quality Assessment: Unifying Structure and Texture Similarity},
  author={Ding, Keyan and Ma, Kede and Wang, Shiqi and Simoncelli, Eero P.},
  journal = {CoRR},
  year={2020},
}

@ARTICLE{vif,
  author={Sheikh, H.R. and Bovik, A.C.},
  journal={IEEE Transactions on Image Processing}, 
  title={Image information and visual quality}, 
  year={2006},
  volume={15},
  number={2},
  pages={430-444},
}

@INPROCEEDINGS{ms-ssim,
  author={Wang, Z. and Simoncelli, E.P. and Bovik, A.C.},
  booktitle={The Thrity-Seventh Asilomar Conference on Signals, Systems \& Computers}, 
  title={Multiscale structural similarity for image quality assessment}, 
  year={2003},
  number={},
  }

@article{scc,
author = {J. Zhou and D. L. Civco and J. A. Silander},
title = {A wavelet transform method to merge Landsat TM and SPOT panchromatic data},
journal = {International Journal of Remote Sensing},
volume = {19},
number = {4},
pages = {743--757},
year = {1998},
}

@INPROCEEDINGS{deblur_diffusion,
  author={Ren, Mengwei and Delbracio, Mauricio and Talebi, Hossein and Gerig, Guido and Milanfar, Peyman},
  booktitle={IEEE/CVF International Conference on Computer Vision (ICCV)}, 
  title={Multiscale Structure Guided Diffusion for Image Deblurring}, 
  year={2023},
  volume={},
  number={},
  pages={10687-10699},
}

@INPROCEEDINGS {deblur_gan,
author = { Kupyn, Orest and Budzan, Volodymyr and Mykhailych, Mykola and Mishkin, Dmytro and Matas, Jiri },
booktitle = {IEEE/CVF Conference on Computer Vision and Pattern Recognition (CVPR) },
title = {{ DeblurGAN: Blind Motion Deblurring Using Conditional Adversarial Networks }},
year = {2018},
volume = {},
ISSN = {},
}

@misc{medsam,
      title={SAM-Med2D}, 
      author={Junlong Cheng and Jin Ye and Zhongying Deng and Jianpin Chen and Tianbin Li and Haoyu Wang and Yanzhou Su and
              Ziyan Huang and Jilong Chen and Lei Jiangand Hui Sun and Junjun He and Shaoting Zhang and Min Zhu and Yu Qiao},
      year={2023},
      eprint={2308.16184},
      archivePrefix={arXiv},
}

@INPROCEEDINGS{sam,
  author={Kirillov, Alexander and Mintun, Eric and Ravi, Nikhila and Mao, Hanzi and Rolland, Chloe and Gustafson, Laura and Xiao, Tete and Whitehead, Spencer and Berg, Alexander C. and Lo, Wan-Yen and Dollár, Piotr and Girshick, Ross},
  booktitle={2023 IEEE/CVF International Conference on Computer Vision (ICCV)}, 
  title={Segment Anything}, 
  year={2023}
}

@inproceedings{haleval,
  title={Hal-eval: A universal and fine-grained hallucination evaluation framework for large vision language models},
  author={Jiang, Chaoya and Jia, Hongrui and Dong, Mengfan and Ye, Wei and Xu, Haiyang and Yan, Ming and Zhang, Ji and Zhang, Shikun},
  booktitle={Proceedings of the 32nd ACM International Conference on Multimedia},
  year={2024}
}

@misc{HaluEval,
  author = {Junyi Li and Xiaoxue Cheng and Wayne Xin Zhao and Jian-Yun Nie and Ji-Rong Wen },
  title = {HaluEval: A Large-Scale Hallucination Evaluation Benchmark for Large Language Models},
  year = {2023},
  journal={arXiv preprint arXiv:2305.11747},
}

@article{hallulens,
      title={HalluLens: LLM Hallucination Benchmark}, 
      author={Yejin Bang and Ziwei Ji and Alan Schelten and Anthony Hartshorn and Tara Fowler and Cheng Zhang and Nicola Cancedda and Pascale Fung},
      year={2025},
      eprint={2504.17550},
}

@misc{autohalluVLM,
      title={AUTOHALLUSION: Automatic Generation of Hallucination Benchmarks for Vision-Language Models}, 
      author={Xiyang Wu and Tianrui Guan and Dianqi Li and Shuaiyi Huang and Xiaoyu Liu and Xijun Wang and Ruiqi Xian and Abhinav Shrivastava and Furong Huang and Jordan Lee Boyd-Graber and Tianyi Zhou and Dinesh Manocha},
      year={2024},
      eprint={2406.10900},
      archivePrefix={arXiv},
      primaryClass={cs.CV},
      url={https://arxiv.org/abs/2406.10900}, 
}

@InProceedings{hallusionbench,
    author    = {Guan, Tianrui and Liu, Fuxiao and Wu, Xiyang and Xian, Ruiqi and Li, Zongxia and Liu, Xiaoyu and Wang, Xijun and Chen, Lichang and Huang, Furong and Yacoob, Yaser and Manocha, Dinesh and Zhou, Tianyi},
    title     = {HallusionBench: An Advanced Diagnostic Suite for Entangled Language Hallucination and Visual Illusion in Large Vision-Language Models},
    booktitle = {Proceedings of the IEEE/CVF Conference on Computer Vision and Pattern Recognition (CVPR)},
    year      = {2024},
}

@inproceedings{pope,
    title = "Evaluating Object Hallucination in Large Vision-Language Models",
    author = "Li, Yifan  and
      Du, Yifan  and
      Zhou, Kun  and
      Wang, Jinpeng  and
      Zhao, Xin  and
      Wen, Ji-Rong",
    booktitle = "Proceedings of the Conference on Empirical Methods in Natural Language Processing",
    year = "2023",
}

@INPROCEEDINGS{haloc,
author = {Park, Eunkyu and Kim, Minyeong and Kim, Gunhee},
title = {HalLoc: Token-level Localization of Hallucinations for Vision Language Models},
  booktitle = {IEEE/CVF Conference on Computer Vision and Pattern Recognition (CVPR)},
  year = {2025},
}

@InProceedings{mvtec,
  author = {Paul Bergmann and Michael Fauser and David Sattlegger and Carsten Steger},
  title = {MVTec AD -- A Comprehensive Real-World Dataset for Unsupervised Anomaly Detection},
  booktitle = {IEEE/CVF Conference on Computer Vision and Pattern Recognition (CVPR)},
  year = {2019},
}

@INPROCEEDINGS{patchcore,
  author={Roth, Karsten and Pemula, Latha and Zepeda, Joaquin and Schölkopf, Bernhard and Brox, Thomas and Gehler, Peter},
  booktitle={Proceedings of the IEEE/CVF Conference on Computer Vision and Pattern Recognition}, 
  title={Towards Total Recall in Industrial Anomaly Detection}, 
  year={2022},
  volume={},
  number={},
  pages={14298-14308}
}

@misc{dinov3,
  title={{DINOv3}},
  author={Sim{\'e}oni, Oriane and Vo, Huy V. and Seitzer, Maximilian and Baldassarre, Federico and Oquab, Maxime and Jose, Cijo and Khalidov, Vasil and Szafraniec, Marc and Yi, Seungeun and Ramamonjisoa, Micha{\"e}l and Massa, Francisco and Haziza, Daniel and Wehrstedt, Luca and Wang, Jianyuan and Darcet, Timoth{\'e}e and Moutakanni, Th{\'e}o and Sentana, Leonel and Roberts, Claire and Vedaldi, Andrea and Tolan, Jamie and Brandt, John and Couprie, Camille and Mairal, Julien and J{\'e}gou, Herv{\'e} and Labatut, Patrick and Bojanowski, Piotr},
  year={2025},
  eprint={2508.10104},
  archivePrefix={arXiv},
}

@inproceedings{resnet50aa,
  title={Making Convolutional Networks Shift-Invariant Again},
  author={Zhang, Richard},
  booktitle={International Conference on Machine Learning (ICML)},
  year={2019}
}

@article{attentionpooling,
    title = "{ABCNN}: Attention-Based Convolutional Neural Network for Modeling Sentence Pairs",
    author = {Yin, Wenpeng  and
      Sch{\"u}tze, Hinrich  and
      Xiang, Bing  and
      Zhou, Bowen},
    journal = "Transactions of the Association for Computational Linguistics",
    volume = "4",
    year = "2016",
    pages = "259--272",
}

@INPROCEEDINGS{imagenet,
  author={Deng, Jia and Dong, Wei and Socher, Richard and Li, Li-Jia and Kai Li and Li Fei-Fei},
  booktitle={IEEE Conference on Computer Vision and Pattern Recognition (CVPR)}, 
  title={ImageNet: A large-scale hierarchical image database}, 
  year={2009},
}

@article{hallucination_aam,
  title={Mitigating Hallucinations in Diffusion Models through Adaptive Attention Modulation},
  author={Oorloff, Trevine and Yacoob, Yaser and Shrivastava, Abhinav},
  journal={arXiv preprint arXiv:2502.16872},
  year={2025}
}

@ARTICLE{mnist,
  author={Lecun, Y. and Bottou, L. and Bengio, Y. and Haffner, P.},
  journal={Proceedings of the IEEE}, 
  title={Gradient-based learning applied to document recognition}, 
  year={1998},
  volume={86},
  number={11},
  pages={2278-2324},
}

@article{hand,
  title={11K Hands: gender recognition and biometric identification using a large dataset of hand images},
  author={Afifi, Mahmoud},
  journal={Multimedia Tools and Applications},
  volume={78},
  number={15},
  pages={20835--20854},
  year={2019},
}
